# Time Series Prediction Using Deep Learning Methods in Healthcare

MOHAMMAD AMIN MORID

Department of Information Systems and Analytics, Leavey School of Business, Santa Clara University, Santa Clara, CA, USA, mmorid@scu.edu

OLIVIA R. LIU SHENG

Department of Operations and Information Systems, David Eccles School of Business, University of Utah, Salt Lake City, UT, USA, olivia.sheng@eccles.utah.edu

JOSEPH DUNBAR

Department of Operations and Information Systems, David Eccles School of Business, University of Utah, Salt Lake City, UT, USA, joseph.dunbar@eccles.utah.edu

Traditional Machine Learning (ML) methods face unique challenges when applied to healthcare predictive analytics. The high-dimensional nature of healthcare data necessitates labor-intensive and time-consuming processes when selecting an appropriate set of features for each new task. Furthermore, ML methods depend heavily on feature engineering to capture the sequential nature of patient data, oftentimes failing to adequately leverage the temporal patterns of medical events and their dependencies. In contrast, recent Deep Learning (DL) methods have shown promising performance for various healthcare prediction tasks by specifically addressing the high-dimensional and temporal challenges of medical data. DL techniques excel at learning useful representations of medical concepts and patient clinical data as well as their nonlinear interactions from high-dimensional raw or minimally-processed healthcare data.

In this paper we systematically reviewed research works that focused on advancing deep neural networks to leverage patient structured time series data for healthcare prediction tasks. To identify relevant studies, we searched MEDLINE, IEEE, Scopus, and ACM digital library for relevant publications through November 4th, 2021. Overall, we found that researchers have contributed to deep time series prediction literature in ten identifiable research streams: DL models, missing value handling, addressing temporal irregularity, patient representation, static data inclusion, attention mechanisms, interpretation, incorporation of medical ontologies, learning strategies, and scalability. This study summarizes research insights from these literature streams, identifies several critical research gaps, and suggests future research opportunities for DL applications using patient time series data.

## 1 INTRODUCTION

As the digital healthcare ecosystem expands, healthcare data is increasingly being recorded within Electronic Health Records (EHR) and Administrative Claims (AC) systems [1,2]. The widespread adoption of these information systems has become popular with government agencies, hospitals, and insurance companies [3,4], capturing data from millions of individuals over many years [5,6]. As a result, physicians, and other medical practitioners are increasingly overwhelmed by the massive amounts of recorded patient data, especially given these professionals' relatively limited access to time, tools, and experience wielding this data on a daily basis [7,8]. This problem has caused machine learning (ML) methods to gain attention within the medical domain, since ML methods effectively use an abundance of available data to extract actionable knowledge, thereby both predicting medical outcomes and enhancing medical decision making [3,9]. Specifically, ML has been utilized in the assessment of early triage, the prediction of physiologic decompensation, the identification of high cost patients, and the characterization of complex, multi-system diseases [10,11], to name a few. Some of these problems, such as early triage assessment, are not new and date back to at least World War I, but the success of ML methods and the concomitant, growing deployment of EHR and AC information systems have sparked broad research interest [4,12].

Despite the swift success of traditional ML in the medical domain, developing effective predictive models remains difficult. Due to the high-dimensional nature of healthcare data, typically only a limited set of appropriate features from among thousands of candidates are selected for each new prediction task, necessitating a labor-intensive and time-consuming process. This often requires the involvement of medical experts to extract, preprocess, and clean data from different sources [13,14]. For example, a recent systematic literature review found that risk prediction models built from EHR data use a median of 27 features from among many thousands of potential variables [15]. Moreover, in order to handle the irregularity and incompleteness prevalent in patient data, traditional ML models are trained using coarse-grain aggregation measures, such as mean and standard deviation, for input features. These depend heavily on manually-crafted features, and they cannot adequately leverage the temporal sequential nature of medical events and their dependencies [16,17]. Another crucial observation is that patient data evolves over time. The sequential nature of medical events, their associated long-term dependencies, and confounding interactions (e.g., disease progression and intervention) offer useful but highly complex information for predicting future medical events [18,19]. Aside from limiting the scalability of traditional predictive models, these complicating factors unavoidably result in imprecise predictions, which can often overwhelm practitioners with false alarms [20,21]. Effective modeling of high-dimensional, temporal medical data can help to improve predictive accuracy and thus increase the adoption of state-of-the-art models in clinical settings [22,23].

Compared with the traditional ML counterpart, deep learning (DL) methods have shown superior performance for various healthcare prediction tasks by addressing the aforementioned high-dimensionality and temporality of medical data [12,16]. These enhanced neural network techniques can learn useful representations of key factors, such as esoteric medical concepts and their interactions, from high-dimensional raw or minimally-processed healthcare data [5,20]. DL models achieve this through repeated sequences of training layers, each employing a large number of simple linear and nonlinear transformations that map inputs to meaningful representations of distinguishable temporal patterns [5,24]. Released from the reliance on experts to specify which manually-crafted features to use, these end-to-end neural net learners have the capability to model data with rich temporal patterns and can encode high-level representations of features as nonlinear combinations of network parameters [25,26].

Not surprisingly, the recent popularity of DL methods has correspondingly increased the number of their associated publications in the healthcare domain [27]. Several studies have reviewed such works from different perspectives. Pandey and Janghel (2019) [28] and Xiao et al. (2018)[29] describe a wide variety of DL models and highlight the challenges of applying them to a healthcare context. Yazhini and Loganathan (2019)[30], Srivastava et al. (2017)[31] and Shamshirband et al. (2021)[32] summarize various applications in which DL models have been successful. Unlike the aforementioned studies, which broadly review DL in various health applications, ranging from genomic analysis to medical imaging, Shickel et al. (2018)[27] exclusively focus on research involving EHR data. They categorize deep EHR learning applications into five categories: information extraction, representation learning, outcome prediction, computational phenotyping, and clinical data de-identification, while describing a theme for each category. Finally, Si et al. (2021)[33] focus on EHR representation learning and investigate their surveyed studies in terms of publication characteristics, which include input data and preprocessing, patient representation, learning approach, and evaluative outcome attributes.

In this paper, we review studies focusing on DL prediction models that leverage patient structured time series data for healthcare prediction tasks from a technical perspective. We do not focus on unstructured patient data, such as images or clinical notes, since DL methods that include natural language processing and unsupervised learning tend to ask research questions that are quite different due to the unstructured nature of the data types. Rather, we summarize the findings of DL researchers for leveraging structured healthcare time series data, of numeric and categorical types, for a target prediction task in terms of the network architecture and learning strategy. Furthermore, we methodically organize how previous researchers have handled the challenging characteristics of healthcare time series data. These characteristics notably include incompleteness, multimodality, irregularity, visit representation, the incorporation of attention mechanisms or medical domain knowledge, outcome interpretation, and scalability. To the best of our knowledge, this is the first review study to investigate these technical characteristics of deep time series prediction in healthcare literature.

## 2 METHOD

### 2.1 Overview

The primary goal of this systematic literature review is to extract and organize the findings from research on structured time series prediction in healthcare using DL approaches, and to subsequently identify related, future research opportunities. Because of their fundamental importance and potential impact, we aimed to address the following review questions:

*1. How are various healthcare data types represented as input for DL methods?*

*2. How do DL methods handle the challenging characteristics of healthcare time series data, including incompleteness, multimodality, and irregularity?*

*3. What DL models are most effective? In what scenarios does one model have advantages over another?*

*4. How can established medical resources help DL methods?*

*5. How can the internal processes of DL outcomes be interpreted to extract credible medical facts?*

*6. To what extent do DL methods developed in limited healthcare settings become scalable to larger healthcare data sources?*

In order to answer these questions, we identify ten core characteristics including medical task, database, input features, preprocessing, patient representation, DL architecture, output temporality, performance, benchmarks, and interpretation for extraction from each study. Section 2.4 elaborates on these ten core characteristics. Additionally, we find that asserted research contributions of the deep time series prediction literature can be classified into the following ten categories: patient representation, missing value handling, DL model, addressing temporal irregularity, attention mechanism, incorporation of medical ontologies, static data inclusion, learning strategy, interpretation, and scalability. Section 3 introduces selected papers exhibiting research contributions in each of these ten categories and further describes their research approaches, hypotheses, and evaluation results. In Section 4, we discuss strengths and weaknesses of the main approaches and identify research gaps based on the same identified ten categories.

## 2.2 Literature search

We searched for eligible articles in MEDLINE, IEEE, Scopus, and the ACM digital library published before February 7th, 2021. In order to show a complete picture of the studies published in 2020, we performed the search and selection process again on November 4th, 2021, and added all studies published in 2020. Our specific search queries for each of these databases can be found in Table S1 of the online supplement.

## 2.3 Inclusion and exclusion criteria

We followed PRISMA guidelines [34] to include English-language, original research studies published in peer-reviewed journals and conference proceedings. Posters and preprints were not included. We specifically selected papers that employed DL methods to leverage structured patient time series data for healthcare prediction tasks. Reviewed works can be broadly classified under the outcome prediction category of Shickel et al. (2018)[27]. We excluded studies based on unstructured data, as well as those lacking key information on the core study characteristics listed in Table 1.

## 2.4 Data extraction

We focused the review of each study to center on the ten identifiable features relating to its problem description, input, methodology, and output. Table 1 provides brief description and explains the main usage of each of these ten characteristics.

Table 1: Core study characteristics

| Feature | | Description | Usage |
|---|---|---|---|
| Medical Task | | Describes the medical time series prediction goal. | Helps to understand if a certain network quality fits a specific task or if it is generalizable to more than one task. |
| Database | | Determines the healthcare data source and scope used for the experiments. | Helps to understand whether the experimental dataset is public or not. Also, since patient data in different countries is recorded with different coding systems, this aids in identifying the adopted coding system. |
| Input | Demographic | Determines if patient demographic data is used as input. | Helps to understand the variety of structured patient data used as input in the study. |
| | Vital Sign | Determines if patient vital signs are used as input. | |
| | Lab Test | Determines if patient lab tests are used as input. | |
| | Procedure Codes | Determines if patient procedure codes are used as input. | |
| | Diagnosis Codes | Determines if patient diagnosis codes are used as input. | |
| | Medication Codes | Determines if patient medication codes are used as input. | |
| | Others | Describes other EHR or AC input features. | |
| Preprocessing | | Describes the windowing and missing value imputation methods. | Helps to understand how data preprocessing affects the outcome. |
| Patient Representation | | Shows the final format of the time series data fed into the DL model. | Helps to identify whether sequence representation or matrix representation has been used to represent patient time series data. |
| DL Architecture | | Shows the DL model architecture used for the time series prediction. | Helps to compare and contrast the learning architectures, and also to identify architectural contributions. |
| Output Temporality | | Determines whether the target is static or dynamic. | Specifies whether the output is the same for a sequence of events or if it changes over time for each event. |
| Performance | | Shows the highest achieved performance based on the primary outcome. | Provides researchers in each learning task with state-of-the-art prediction performance. |
| Benchmark | | Lists the models used as a baseline for comparison. | Identifies traditional ML or DL models that are outperformed by the proposed model. |
| Interpretation | | Shows the methods used for DL model interpretation. | Aids in understanding how a DL "black-box" model has been interpreted. |

## 2.5 Data analysis

Each selected study in the systematic review has either proposed a technical contribution to advancing the methods for a deep time series prediction pipeline or adopted an extant method for a new healthcare application to make a domain contribution. The focus of this systematic review is summarizing the findings of the former research stream with technical contributions and identifying the associated research gaps. Nevertheless, we also briefly summarize and discuss articles with domain contributions.

Based on the technical contributions noted in the included studies, we classify identifiable contributions into one of ten categories: *patient representation, missing value handling, DL models, addressing temporal irregularity, attention mechanisms, incorporation of medical ontologies, static data inclusion, learning strategy, interpretation, and scalability*. For each given category, Section 3 summarizes deep patient time series learning approaches identified in the reviewed studies. Section 4 compares the strengths and weaknesses of these DL techniques and the associated future research opportunities.

## 3 RESULTS

Our literature search initially resulted in 1,524 studies, with 511 of them being duplicates (i.e., indexed in multiple databases). The remaining 1,014 works underwent a title and abstract screening. Following our exclusion criteria, 621 studies were excluded. Out of these 621 omitted studies, 74 did not use EHR or AC data, 81 did not use multivariate temporal data, 171 did not use DL methods for their prediction tasks, and 295 studies were based on unstructured data, such as images, clinical notes, or sensor data. The remaining 393 papers were then selected for a full-text review, and we subsequently removed 316 additional papers because they lacked one or more of the core study characteristics listed in Table 1. Specifically, 64 of the removed papers did not provide distinctive input features (e.g., medical code types), 99 did not have patient representation (e.g., embedding vector creation), 129 did not sufficiently describe their DL network architectures (e.g., RNN network type), and 24 did not specify their output temporality (i.e., static or dynamic) designs. Below, Figure 1 summarizes the article extraction procedure, and Figure 2 shows the distribution of the 77 included studies based on their publication year. A majority of the studies (77%) were published after 2018, signaling a recent surge in interest among researchers for DL models applied to healthcare prediction tasks.

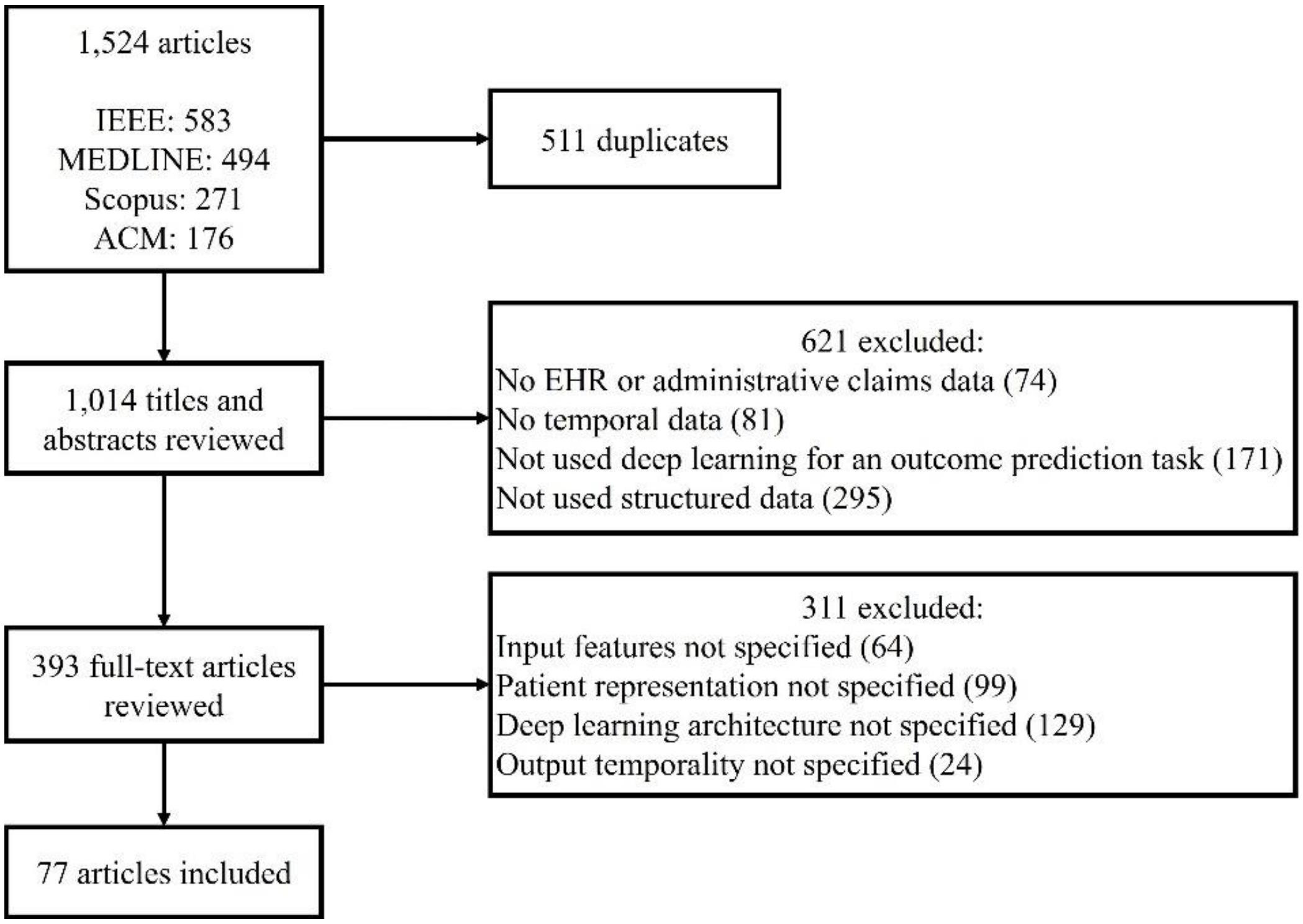

Figure 1: Inclusion flow of the systematic review.

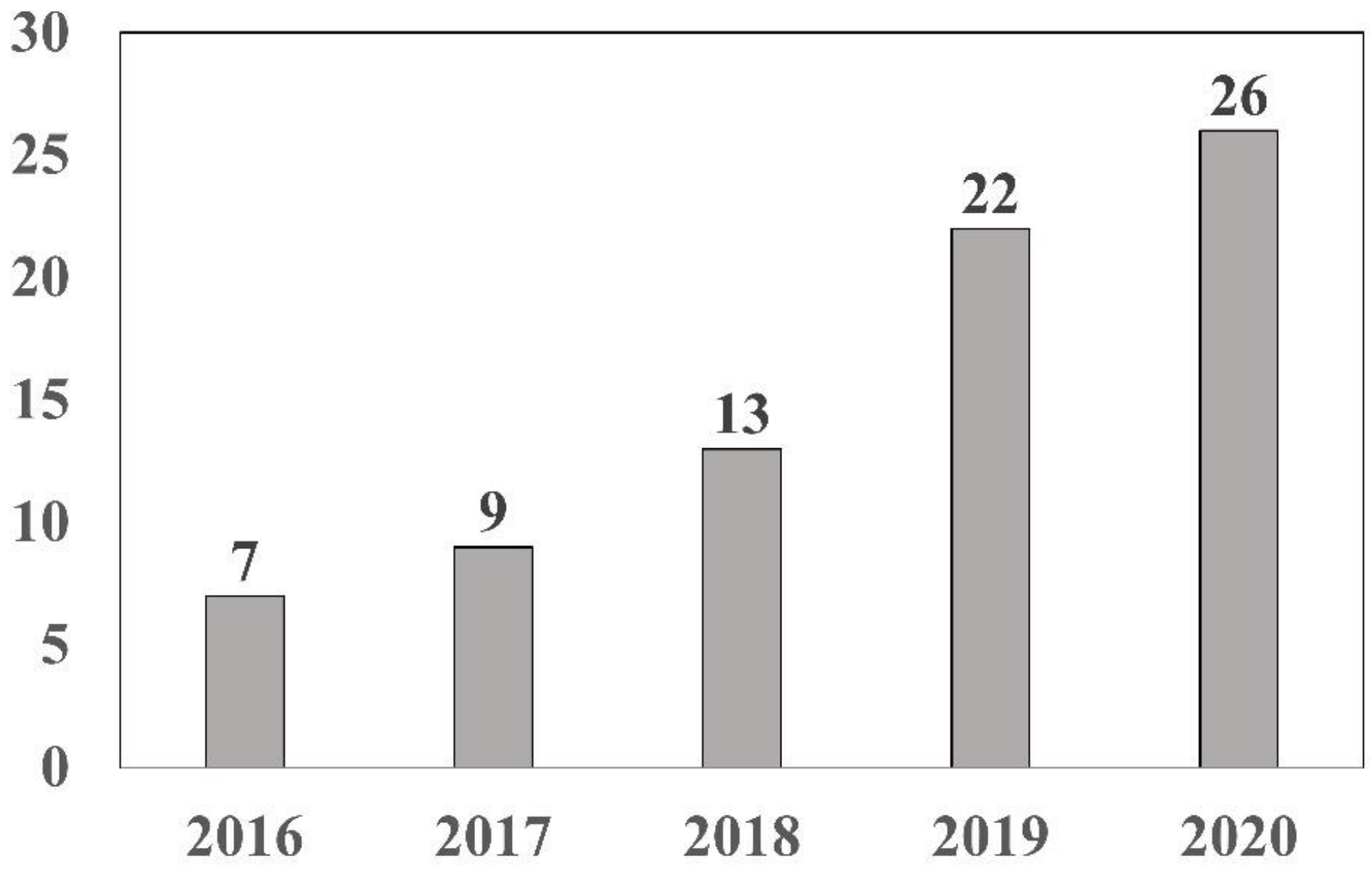

Figure 2: Number of publications per year.

Table 2 lists the included studies by prediction task. Note that mortality, heart failure, readmission, and patient next-visit diagnosis predictions are the most studied prediction tasks, and a publicly available online dataset, the Medical Information Mart for Intensive Care (MIMIC) [35], is the most popular data source for the studies. A complete list of the included studies and their characteristics as delineated in Table 1 is available in the online supplement (Tables S2 and S3).

Table 2: List of reviewed studies per prediction task.

| Prediction Task | Reference |
|---|---|
| Mortality | Che et al. (2018)[36], Sun et al. (2019)[9], Yu et al. (2020)[37], Ge et al. (2018)[38], Caicedo-Torres et al. (2019)[39], Sha et al. (2017)[2], Harutyunyan et al. (2019)[12], Rajkomar et a. (2018)[13], Zhang et al. (2020)[40], Shickel et al. (2019)[41], Purushotham et al. (2018)[42], Gupta et al. (2020)[43], Baker et al. (2020)[44], Yu et al. (2020)[45] |
| Heart Failure | Cheng et al. (2016)[18], Choi et al. (2017)[46], Yin et al. (2019)[47], Wang et al. (2018)[48], Ju et al. (2020)[49], Bekhet et al. (2019)[50], Choi et al. (2016)[22], Maragatham et al. (2019)[51], Zhang et al. (2019)[52], Choi et al. (2017)[53], Ma et al. (2018)[54], Choi et al. (2018)[55], Solares et al. (2020)[56] |
| Readmission | Zhang et al. (2018)[57], Wang et al. (2018)[58], Lin et al. (2019)[59], Barbieri et al. (2020)[60], Min et al. (2019)[1], Ashfaq et al. (2019)[61], Rajkomar et a. (2018)[13], Reddy et al. (2018)[62], Nguyen et al. (2017)[63], Zhang et al. (2020)[40], Solares et al. (2020)[56] |
| Next Visit Diagnosis | Lipton et al. (2015)[64], Choi et al. (2016)[7], Pham et al. (2016)[65], Wang et al. (2019)[20], Yang et al. (2019)[66], Guo et al. (2019)[67], Wang et al. (2019)[68], Ma et al. (2017)[69], Ma et al. (2018)[70], Harutyunyan et al. (2019)[12], Pham et al. (2017)[71], Lee et al. (2019)[72], Rajkomar et a. (2018)[13], Lee et al. (2020) [73], Choi et al. (2017)[53], Purushotham et al. (2018)[42], Gupta et al. (2020)[43], Lipton et al. (2016)[74], Bai et al. (2018)[75], Liu et al. (2020)[76], Zhang et al. (2020)[77], Qiao et al. (2020)[78] |
| Cardiovascular Disease | Che et al. (2018)[36], Park et al. (2019)[79], An et al. (2019)[80], Duan et al. (2019)[81], Park et al. (2018)[82] |
| Length of Stay | Che et al. (2018)[36], Harutyunyan et al. (2019)[12], Rajkomar et al. (2018)[13], Zhang et al. (2020)[40], Purushotham et al. (2018)[42] |
| Sepsis Shock | Zhang et al. (2017)[83], Zhang et al. (2019)[84], Wickramaratne et al. (2020)[85], Svenson et al. (2020)[86], Fagerström et al. (2019)[87] |
| Hypertension | Mohammadi et al. (2019)[88], Ye et al. (2020) [89] |
| Decompensation | Harutyunyan et al. (2019)[12], Purushotham et al. (2018)[42], Thorsen-Meyer et al. (2020)[90] |
| Illness Severity | Chen et al. (2018)[17], Zheng et al. (2017)[91], Sue et al. (2017)[92] |
| Acute Kidney Injury | Tomasev et al. (2019)[93] |
| Joint Replacement Surgery Risk | Qiu et al. (2019)[94] |
| Post-Stroke Pneumonia | Ge et al. (2019)[95] |
| Renal Disease | Razavian et al. (2016)[96] |
| Adverse Drug Event | Rebane et al. (2019)[97] |
| Cost | Morid et al. (2020)[98] |
| Chronic Obstructive Pulmonary Disorder | Cheng et al. (2016)[18] |
| Kidney Transplantation Endpoint | Esteban et al. (2016)[3] |
| Surgery Recovery | Che et al. (2018)[36] |
| Diabetes | Ju et al. (2020)[49] |
| Asthma | Xiang et al. (2020)[99] |
| Neonatal Encephalopathy | Gao et al. (2019)[100] |

After reviewing the included studies, we found that the asserted contributions of researchers within the deep time series prediction literature can be distinguished and classified under the following ten categories: (1) patient representation, (2) missing value handling, (3) DL model, (4) addressing temporal irregularities, (5) attention mechanism, (6) incorporation of medical ontologies, (7) static data inclusion, (8) learning strategy, (9) interpretation strategies, and (10) scalability. The rest of Section 3 devotes one subsection for each of these categories to describe the associated findings by category. Figure 3 below gives a general overview of the focal approaches adopted by the included studies.

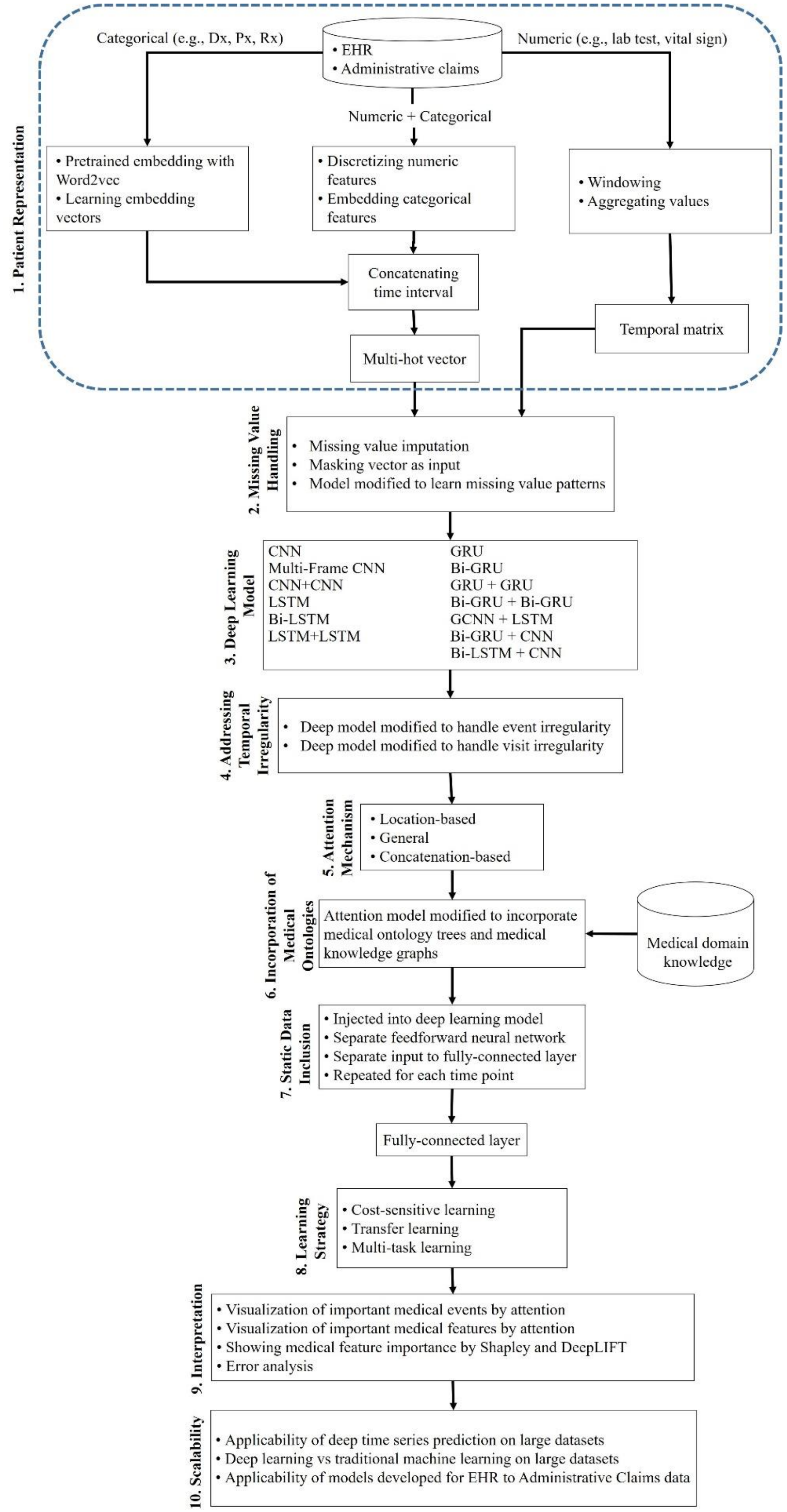

Figure 3: Summary of deep patient time series prediction designs

### 3.1 Patient representation

Patient representations employed for deep time series prediction in healthcare can broadly be classified into one of two categories: sequence representation and matrix representation [1]. In the former approach, each patient is represented as a sequence of medical event codes (e.g., diagnosis code, procedure code, or medication code), and the additional input may or may not include the time interval between the events (Section 3.3). Since a complete list of medical codes is generally quite long, various embedding techniques are commonly used to shorten it or combine similar medical codes with comparable values. In the latter approach, each patient is represented as a longitudinal matrix, where columns correspond to different medical events, and rows correspond to regular time intervals. As a result, a cell in a patient matrix provides the code of the patient's medical or claims event at a particular time point. Zhang et al. (2018)[57] followed a hybrid approach which splits the overall patient sequence of visits into multiple subsequences of equal length, and then embeds the medical codes in each subsequence as a multi-hot vector.

As seen in Table S3, sequence representation is a slightly more prevalent approach employed by researchers (57%). Generally, for prediction tasks with numeric inputs, such as lab tests or vital signs, sequence representation is more commonly used, and for those with categorical inputs, like diagnosis codes or procedure codes, matrix representation is the trend. Nevertheless, there are some exceptions. Rajkomar et al. (2018)[13] converted patient lab test results from numeric values to categories by assigning a unique token to each lab test name, value, and unit (e.g., "Hemoglobin 12 g/dL") for predicting mortality, length-of-stay, and readmission in Intensive Care Units (ICUs). Ashfaq et al. (2019)[61] included the lab test code with a value if the value was designated to be abnormal (determined according to medical domain knowledge), in addition to the typical inclusion of diagnosis and procedure codes. Several research groups [72,80,89] converted numerical lab test results into predesigned categories by encoding them as either missing, low, normal, or high when predicting hypertension and the associated onset of high-risk cardiovascular states. Similarly, Barbieri et al. (2020)[60] transformed vital signs into OASIS severity scores, and then discretized these scores into categories of low, normal, and high. Of note, a singular study observed the superiority of matrix representation over sequence representation for readmission prediction of Chronic Obstructive Pulmonary Disease (COPD) patients using a large AC database [1]. This study and other matrix representations [44,57,96] found that integrating coarse time granularities such as weekly or monthly, rather than finer time granularity measures can improve performance. This study also compared various embedding techniques, and the authors found no significant differences in their results. Finally, Qiao et al. (2020)[78] summarized each numerical time series in terms of temporal measures such as their self-correlation structure, data distribution, entropy, and stationarity. They found that these measures can improve the interpretability of the extracted temporal features without degrading prediction performance.

For embedding medical events in the sequence representation, a commonly observed technique was to augment the neural network with an embedding layer that can learn effective medical code representations. This technique has benefitted the prediction of hospital readmission [58], patient next-visit diagnosis [66], and the onset of vascular diseases [82]. Another event embedding technique has been to use a pre-trained embedding layer via probabilistic methods, especially word2vec [101] and Skip-Gram [102], which have shown promising results for predicting an assortment of healthcare outcomes, such as patient next-visit diagnosis [7], heart failure [46,51], and hospital readmission [57]. Choi et al. (2016)[7] demonstrated that pre-trained embedding layers can outperform trainable layers by a 2% margin in recall for the next-visit diagnosis prediction problem. Instead of relying on individual medical codes for the next-visit diagnosis problem, several studies grouped medical codes using the first three digits of each diagnosis code, and other works implemented Clinical Classification Software (CCS) [103] to obtain groupings of medical codes [68,73]. On the other hand, Maragatham et al. (2019)[51] observed that pre-trained embedding layers can outperform medical group coding methods by a 1.5% margin in Area Under the Curve (AUC) for heart failure prediction.

Finally, Min et al. (2019)[1] showed that, independent of the embedding approach, patient matrix representation generally outperformed sequence representation.

### 3.2 Missing value handling

Missing value imputation using methods such as zero [3,40], median [58], forward-backward [64,66] and domain-knowledge by experts [12,38] has been the most common approach for handling missing values in patient time-series data. Lipton et al. (2016)[74] was the first study that used a masking vector to utilize the availability of values as a separate input to predict discharge diagnosis. Other studies adopted the same approach for predicting readmission [59], acute kidney injury [93], ICU mortality [37], and length-of-stay [12]. Lastly, Che et al. (2018) [36] utilized missing patterns as input for predicting mortality, length-of-stay, surgery recovery, and cardiac condition. Their approach outperformed the masking vector technique by approximately 2% margin in AUC.

### 3.3 Deep learning model

Table 3 shows the summary of model architectures adopted to learn a deep patient time series prediction model for each included study. Recurrent Neural Networks (RNN) and their modern variants, including Long Short-Term Memory (LSTM) and Gated Recurrent Units (GRU), were by far the most frequently used models (84%). A few studies compared the GRU variant against the LSTM architecture. Overall, GRU achieved around 1% advantage in AUC metrics over LSTM for predicting heart failure [47], kidney transplantation endpoint [3], mortality in the ICU [36], and readmission prediction of chronic disease patients [1]. However, for predicting the diagnosis code group of a patient's next admission to the ICU [68], sepsis shock [83], and hypertension [89], researchers did not find significant differences between these two advanced RNN model types. Additionally, bidirectional variants of GRU and LSTM—so-called Bi-GRU and Bi-LSTM—consistently outperformed their unidirectional counterparts for predicting hospital readmission [57], diagnosis at hospital discharge [66], patient next-visit diagnosis [67,69,75], adverse cardiac events [81], readmission after ICU discharge [59,60], mortality in hospital [2,45], length-of-stay in hospital [12], sepsis [85], and heart failure [54]. While the majority of studies (63%) employed single-layered RNN, many other works used multi-layered RNN models with GRU [7,48], LSTM [40,64,68,74], and Bi-GRU [2,67,82]. However, despite the numerous studies employing these methods and their variants, multi-layered GRU is the only architecture that has been experimentally compared to its single-layered counterpart for the patient next-visit diagnosis [7] and heart failure prediction tasks [48]. Alternatively, researchers have extensively explored training separate network layers with the architectures of LSTM [12,38], Bi-LSTM [77] and GRU layers [17] for each feature. These channel-like architectures per feature were reported as being more successful than the simpler RNN models. Finally, for tasks such as predicting in-hospital mortality or hospital discharge diagnosis code, some RNN models were supervised to make assessments at each time step [12,64,74], a procedure known as *target replication*. Their successes provided evidence that it can be more effective to repeatedly make a prediction at multiple time points than merely performing supervised learning for the last time-stamped entry.

Table 3: Deep learning model architectures of the reviewed studies

| Model | Reference |
|---|---|
| CNN | Caicedo-Torres et al. (2019)[39], Nguyen et al. (2017)[63] |
| Multi-Frame CNN | Cheng et al. (2016)[18], Ju et al. (2020)[49] |
| CNN+CNN | Razavian et al. (2016)[96], Wang et al. (2018)[58], Morid et al. (2020)[98] |
| LSTM | Pham et al. (2016)[65] [71], Zhang et al. (2017)[83], Rajkomar et a. (2018)[13], Wang et al. (2019)[20], Gao et al. (2019)[100], Qiu et al. (2019)[94], Mohammadi et al. (2019)[88], Park et al. (2019)[79], Ashfaq et al. (2019)[61], Maragatham et al. (2019)[51], Yu et al. (2020)[37], Ye et al. (2020) [89], Maragatham et al. (2019)[51], Reddy et al. (2018)[62], Lee et al. (2019)[72], Zhang et al. (2019)[84], Xiang et al. (2020)[99], Thorsen-Meyer et al. (2020)[90] |
| Bi-LSTM | Yang et al. (2019)[66], Ye et al. (2020) [89], Bai et al. (2018)[75], Duan et al. (2019)[81], Yu et al. (2020)[45] |
| LSTM+LSTM | Lipton et al. (2015)[64], Lipton et al. (2016)[74], Yin et al. (2019)[47], Wang et al. (2019)[68], Zhang et al. (2020)[40], Fagerström et al. (2019)[87] |
| GRU | Esteban et al. (2016)[3], Choi et al. (2017)[46], Zheng et al. (2017)[91], Choi et al. (2017)[53], Choi et al. (2016)[22], Che et al. (2018)[36], Ma et al. (2018)[70], Purushotham et al. (2018)[42], Tomasev et al. (2019)[93], Bekhet et al. (2019)[50], Min et al. (2019)[1], Shickel et al. (2019)[41], Solares et al. (2020)[56], Choi et al. (2018)[55], Rebane et al. (2019)[97], Ge et al. (2019)[95], Sue et al. (2017)[92], Liu et al. (2020)[76], Zhang et al. (2020)[77] |
| Bi-GRU | Ma et al. (2017)[69], Wickramaratne et al. (2020)[85], Zhang et al. (2018)[57], Barbieri et al. (2020)[60], Sun et al. (2019)[9], Qiao et al. (2020)[78] |
| GRU + GRU | Choi et al. (2016)[7], Wang et al. (2018)[48], Gupta et al. (2020)[43] |
| Bi-GRU + Bi-GRU | Sha et al. (2017)[2], Park et al. (2018)[82], Guo et al. (2019)[67] (concurrent) |
| GCNN + LSTM | Lee et al. (2020) [73] |
| Bi-GRU + CNN | Ma et al. (2018)[54] |
| Bi-LSTM + CNN | Lin et al. (2019)[59], Baker et al. (2020)[44] |
| One RNN per Feature or Feature Type | Ge et al. (2018)[38], Harutyunyan et al. (2019)[12], An et al. (2019)[80], Chen et al. (2018)[17], Svenson et al. (2020)[86] |

Several studies, particularly those from when deep time series prediction within the healthcare domain was in its nascency, utilized convolutional neural network (CNN) models for prediction tasks without benchmarking against other types of DL models [18,39,58]. These early CNN models have been consistently outperformed by recently-developed RNN models for predicting heart failure [49,52], readmission of patients diagnosed with chronic disease [1], in-hospital mortality [40], diabetes [49], readmission after ICU discharge [40,59], and joint replacement surgery risk [94]. Nevertheless, Cheng et al. (2016)[18] showed that temporal slow fusion can enhance CNN performance, and Ju et al. (2020)[49] suggested using 3D-CNN and spatial pyramid pooling for outperforming RNN models for heart failure and diabetes prediction tasks. Alternatively, hybrid deployments of CNN/RNN models have been successful in outperforming pure CNN or RNN models for predicting readmission after ICU discharge [59], patient next-visit diagnosis [73], mortality [44], and heart failure [54].

### 3.4 Addressing temporal irregularity

Two types of temporal irregularities, visit and feature, generally exist in patient data. Visit irregularity indicates that the time interval between visits can vary for the same patient over time. Feature irregularity occurs when different features belonging to the same patient for the same visit are recorded at various time points and frequencies.

Che et al. (2016)[7] was the first study to make use of the time interval between patient visits as a separate input to a DL model for the patient next-visit diagnosis prediction task. This approach also proved to be efficacious in predicting heart failure [46], vascular diseases [82], hospital mortality [13] and hospital readmission [13]. Yin et al. (2019)[47] used a sinusoidal transformation of time interval for assessing heart failure. Additionally, Pham et al. (2016)[65] and Wang et al. (2019)[20] modified the internal mechanisms of the LSTM architecture to handle visit irregularity by giving higher weights to recent visits. Their proposed modifications outperformed traditional LSTM architectures by 3% in AUC for the highly frequent benchmarking task of predicting the diagnosis code group of a patient's next visit.

Certain studies hypothesized that handling feature irregularity is more effective than handling visit irregularity [60,91]. Zheng et al. (2017)[91] also modified GRU memory cell learning processes to extract different decay patterns for each input feature for predicting the Alzheimer's severity score in half-a-year. Their results demonstrated that capturing feature and visit irregularity decreases the mean squared error (MSE) by up to 5% compared to models that capture visit irregularity only. Barbieri et al. (2020)[60] and Liu et al. (2020)[76] used a similar approach when predicting readmission to ICU and for generating relevant medications from billing codes.

### 3.5 Attention mechanism

Attention mechanisms, originally inspired by the visual attention system found in human physiology, have recently become quite popular among many domains, including deep time series prediction for healthcare [57]. The core underlying idea is that patient visits and their associated medical events should not carry an identical weight during the inference process. Rather, they are contingent on their relative importance for the prediction task at-hand.

Most commonly, attention mechanisms initially assign a unique weight for each visit or each medical event, and subsequently optimize these weight parameters during network backpropagation [2,13,22,37]. Also called location-based attention [69], this strategy has been incorporated into a variety of RNN networks and learning tasks, such as GRU for heart failure [22], Bi-GRU for mortality [51], as well as LSTM for hospital readmission, diagnosis, length-of-stay [13], and asthma exacerbation [99]. Other commonly used attention mechanisms include a concatenation-based attention device which has been employed for hospital readmission [60] as well as next-visit diagnosis prediction [69], and general attention models which are used primarily for hospital readmission [57] and mortality prediction [41]. Ma et al. (2017)[69] benchmarked these three attention mechanisms for predicting medical codes by using a large AC database, and Sue et al. (2017)[92] performed a similar benchmarking procedure for illness severity score prediction on EHR data. Both studies reported location-based attention as optimal.

With few exceptions, studies employing an attention mechanism tended not to report any differential prediction performance improvements enabled by attention. Those few studies which did distinguish a particular performance improvement reported that location-based attention mechanisms improved patient next-visit diagnosis by 4% in AUC [65], increased hospital readmission F1-Score by 2.4%, and also saw a 13% boost in F1-Score for mortality prediction [2]. Zhang et al. (2018)[57] was the sole work reporting contributions of visit-level attention and medical code attention separately for hospital readmission, observing that each technique provided an approximate 4% increase in F2-Score. An innovative study by Guo et al. (2019)[67] argued that all medical codes should not go through the same weight allocation path during attention calculation. Instead, they proposed a crossover attention model with distinct bidirectional GRUs and attention weights for both diagnosis and medication codes. On the whole, we found that most studies utilized attention mechanisms to improve the interpretability of their proposed DL models by highlighting important visits or medical codes, at either a patient or population level. Section 3.9 further elaborates on patient and population level properties.

### 3.6 Incorporation of medical ontologies

Another facet of these research streams was the incorporation of medical domain knowledge into DL models to enhance their prediction performance. Standard Clinical Classification Software (CCS) has the ability to establish a hierarchy of various medical concepts in the form of successive parent-child relationships. Based on this concept, Choi et al. (2017)[53] employed CCS to create a medical ontology tree for use in a network embedding layer. These encoded medical ontologies were better able to represent abstract medical concepts when predicting heart failure. Zhang et al. (2020)[77] later enhanced this initial ontological strategy by considering more than one parent for each node and also by providing an ordered set of ancestors for each medical concept. Separately, Ma et al. (2018)[70] showed that medical ontology trees can be leveraged when calculating attention weights in GRU models, achieving a 3% accuracy increase over [53] for the same prediction task. Following this, Yin et al. (2019)[47] demonstrated that causal medical knowledge graphs like KnowLife[104], which contain both “cause” and “is-caused-by” relationships between diseases, outperform both [53] and [70] with an approximate 2% AUC margin for heart failure prediction. Wang et al. (2019)[20], on the other hand, enhanced SkipGram embeddings by adding n-gram tokens from medical concept information, such as disease or drug name, to EHR data. These embedded tokens captured ancestral information for a medical concept similar to ontology trees, and they were applied to the patient next-visit diagnosis task.

### 3.7 Static data inclusion

RNN networks are particularly apt at learning from sequential data, though leveraging static data into these types of models has been challenging. The hybrid combination of static along with temporal input is particularly important in a healthcare context, since static features like patient demographic information and prior history can be essential for achieving accurate predictions. Appending patient static data to the input of a final fully-connected layer has been the most common approach for integrating these features. It has been applied to hospital readmission [57,58], length-of-stay [40] and mortality [38,40] tasks. Alternatively, Esteban et al. (2016)[3] fed 342 static features into an entirely independent feedforward neural network before combining the output with temporal data in a typical GRU layer for learning kidney transplant endpoints. Other studies also adopted this approach for predicting mortality [42], phenotyping [42], length-of-stay [42], and the risk of cardiovascular diseases [80]. Moreover, Pham et al. (2016)[65] modified the internal processes of LSTM networks to specifically incorporate the effects of unplanned hospital admissions, which involve higher risks than planned admissions. They employed this approach for predicting patient next-visit diagnosis codes in mental health and diabetes cohorts. Finally, Maragatham et al. (2019)[51] converted static data into a temporal format by repeating it as input to every time point. Together, they used static demographic data, vascular risk factors, and a scored assessment of nursing levels for heart failure prediction. We found no study comparing the aforementioned static data inclusion methods against solid benchmarks.

### 3.8 Learning strategy

We identified three principal learning strategies which differ from the basic supervised learning scenario: (1) cost-sensitive learning, (2) multi-task learning, and (3) transfer learning. When handling imbalanced datasets, cost-sensitive learning has frequently been implemented by modifying the cross entropy loss function [58,61,93,100]. In particular, two studies convincingly demonstrated the performance improvement achieved by cost-sensitive learning. Gao et al. (2019)[100] found a 3.7% AUC increase for neonatal encephalopathy prediction and Ashfaq et al. (2019)[61] observed a 4% increase for the hospital readmission task. The latter study further calculated cost-saving outcomes by estimating the potential annual cost savings if an intervention is selectively offered to patients at high risk for readmission. Instead, multi-task learning was implemented to jointly predict mortality, length-of-stay, and phenotyping with LSTM [13,40], Bi-LSTM [12], and GRU [42] architectures. Harutyunyan et al. (2019)[12] was a seminal study that reported a significant contribution of multi-task learning over state-of-the-art traditional learning, with a solid 2% increase in AUC. Lastly, transfer learning, originally used as a benchmark evaluation by [36], was recently adopted by Gupta et al. (2020)[43] to study both task adaptation and domain adaptation utilizing a non-healthcare model, TimeNet. They found that domain adaptation outperforms task adaptation when the data size is small, but otherwise task adaptation is superior. Moreover, they found that, for task adaption on medium-sized data, fine-tuning is a better approach than learning from scratch with feature extraction.

### 3.9 Interpretation

By far the most common DL interpretation method is to show visualized examples of selected patient records to highlight which visits and medical codes most influence the prediction task [2,13,22,41,47,49,54,57,60,66,67,69,75,82,95,97]. Specific contributions by feature are extracted from the calculated weight parameters of an attention mechanism (Section 3.6). Visualizations can also be implemented through a global average pooling layer [65,82] or a one-sided convolution layer within the neural network [57]. Another interpretation approach is to report the top medical codes with the highest attention weights for all patients together [2] or for different patient groups by disease [47,57,69,80]. Specifically, Nguyen et al. (2017)[63] extracted the most frequent patterns in medical codes by disease type, and Caicedo-Torres et al. (2019)[39] identified important temporal features for mortality prediction using both DeepLIFT [105] and Shapley [106] values. The technique of using Shapley values for interpretation was also employed for continuous mortality prediction within the ICU setting [90]. Finally, Choi et al. (2017)[46] performed error analysis on false positive and false negative predictions to differentiate the contexts in which their DL models are more or less accurate.

### 3.10 Scalability

While most review studies evaluated their proposed models on a single dataset—usually a publicly available resource such as MIMIC and its updates [35]—certain studies focused on assessing the scalability of their models to a wider variety of data. Bekhet et al. (2019)[50] evaluated one of the most popular deep time series prediction models with two GRU layers, called RETAIN which was first proposed in study [22], on a collection of ten hospital EHR datasets for heart failure prediction. Overall, they achieved a similar AUC compared to the original study, though a higher dimensionality did further improve prediction performance. Using the same RETAIN model, Solares et al. (2020)[56] conducted a scalability study on approximately four million patients in the UK National Health Service, and they reported an identical observation to [49]. Another large dataset was explored by Rajkomar et al. (2018)[13], who demonstrated the power of LSTM models on a variety of healthcare prediction tasks for 216,000 hospitalizations involving 114,000 unique patients. Finally, we found a singular study [1] investigating the scalability of deep time series prediction methods for AC data, as opposed to EHR sequences. Min et al. (2019) [1] observed that DL models are effective for readmission prediction with patient EHR data, but they tend not to be superior to traditional ML models using AC data.

Studies on the MIMIC database have consistently used the same 17 features in the dataset which have a low missing rate [107]. To address dimensional scalability, Purushotham et al. (2018)[42] attempted using as many as 136 features for mortality, length-of-stay, and phenotype prediction with a standard GRU architecture. Compared to an ensemble model constructed from several traditional ML models, they found that, for lower-dimensional data, traditional ML performance is comparable to DL performance, while for high-dimensional data, DL's advantage is more pronounced. On a similar note, Min et al. (2019)[1] evaluated a GRU architecture against traditional supervised learning methods on around 103 million medical claims and 17 million pharmacy claims for 111,000 patients. Again, they found that strong traditional supervised ML techniques have a comparable performance to that of their DL competitors.

## 4 DISCUSSION

### 4.1 Patient representation

Out of the commonly used sequential and matrix patient representations, prediction tasks with predominantly numeric inputs such as lab tests and vital signs, often rely on sequence representations, whereas those studies utilizing mainly categorical inputs, like diagnosis codes or procedure codes, commonly incorporate a matrix representation. Other than a lone study [1] that documented the superiority of the matrix approach on AC data, we found no consistent comparison between these two approaches in our systematic review. Also, while a coarse grain abstraction has not been suggested for each of these approaches, changing the granularity level to find the optimal level would be highly suggested to further ascertain their respective efficacy. The rationale is that the sparsity of temporal patient data is typically high, and considering every individual visit for an embedded patient representation may not be the optimal approach when factoring in the corresponding increase in computational complexity.

In order to combine numeric and categorical input features, researchers have generally employed three distinct methods. One method involves converting patient numeric quantities to categorical ones by assigning a unique token to each measure. Thus, each specific lab test code, value, and unit will have its own identifying marker. Using a second method, researchers encode numeric measures with clinically meaningful names, such as *missing*, *low*, *high*, *normal*, and *abnormal*. A third alternative requires the conversion of numeric measures to severity scores, in order to discretize them into *low*, *normal*, and *high* categories. The second approach was quite common in our selected studies, likely due to its implementation simplicity and effectiveness for a wide variety of clinical healthcare applications. We therefore report it to be the most dominant strategy for combining numeric and categorical inputs for deep time series prediction tasks.

When embedding medical events into a sequence representation, we again found three prevalent techniques. Using the first technique, researchers commonly added a separate embedding layer, prefacing the bulwark of the recurrent network, in order to optimize medical code representation. Alternatively, pre-trained embedding layers with established methods such as word2vec were adopted in lieu of learning embeddings from scratch. Lastly, researchers often utilized medical code groups instead of the atomized medical code. Among the three practices, pre-trained embedding layers have consistently outperformed naïve embedding layers and medical code groupings for EHR data, while no significant difference in model performance has been observed for AC data. In addition, researchers have shown that temporal matrix representation is the most effective approach for AC data. The rationale is that the temporal granularity of EHR data is usually at the level of an *hour* or even *minute*, while the granularity of AC data is at the *day* level. As a result, the order of medical codes within a day is ordinarily lost for the embedding algorithms such as word2vec. Combining our findings, a sequence representation with a pre-trained embedding layer is highly recommended for learning tasks on EHR data, while a matrix representation seems to be more effective for AC data.

Several important gaps exist regarding the specific representation of longitudinal patient data. Sequence and matrix methodologies should be compared in a sufficient variety of healthcare settings for EHR data. If extensive comparisons could confirm the relative performance of matrix representation, then it would further enhance its desirability, as it is easier to implement and has a faster runtime than sequences of EHR codes. Moreover, to improve patient similarity measures, researchers should analyze the effect of different representation approaches under various DL model architectures. Lastly, we found that very few reviewed studies included both numerical and categorical measures as feature input. A superior approach which synergistically combines their relative strengths has not yet been sufficiently studied, and thus requires the attention of future research. Further investigation of novel DL architectures with a variety of possible input measures is therefore recommended.

### 4.2 Missing value handling

The most common missing value handling approach found in the deep time series prediction literature was imputation by pre-determined measures, such as zero or the median—also a common practice in non-healthcare domains [108]. However, missing values in healthcare data typically do not occur at random as they can reflect specific decisions by caregivers [74]. These missing values thus represent *informative missingness*, providing rich information about target labels [36]. In order to capture this correspondence, researchers have implemented two primary approaches. The first approach involves creating a binary (masking) vector for each temporal variable, indicating the availability of data at each time point. This approach has been evaluated in various applications, and it seems to be an effective way of handling missing values. Second, missing patterns can be learned by directly training the imputation value as a function of either the latest observation or the empirical mean prior to variable observations. This latter approach is more effective when there is a high missing rate and a high correlation between missing values and the target variable. For instance, Che et al. (2018) [36] found that learning missing values was more effective when the average Pearson correlation between lab tests with a high rate of missingness and the dependent variable, mortality, was above 0.5. Despite this, since masking vectors have been evaluated on a wider variety of healthcare applications, and with different degrees of missingness, they should remain as the suggested missing value handling strategy for deep time series prediction.

Interestingly, there was no study assessing the differential impact of missingness for individual features on a given learning task. The identification of features whose exclusion or missingness most harms the prediction process informs practitioners about how to focus their data collection and imputation strategies. Furthermore, while *informative missingness* applies to many temporal features, *missing-at-random* can still be the case for other feature types. As a direction for future study, we recommend a comprehensive analysis of potential sources of missingness, for each feature and its type, along with assistance from domain experts. This would better inform a missing value handling approach within the healthcare domain, and as a consequence, enhance prediction performance accordingly.

### 4.3 Deep learning models

Rooted in their ability to efficiently represent sequential data and extract its temporal patterns [64], RNN-based DL models and their variants were found to be the most prevalent architecture for deep time series prediction on healthcare data. Patient data naturally has a sequential nature, where hospital visits or medical events occur chronologically. Lab test orders or vital sign records, for example, take place at specific timestamps during a hospital visit. However, vanilla RNN architectures are not sophisticated enough to sufficiently capture temporal dependencies when EHR sequences are relatively long, due to the vanishing gradient problem [109]. To address this issue, LSTM and GRU recurrent networks, with their memory cells and elaborate gating mechanisms, have been habitually employed by researchers, with improved outcomes on a variety of healthcare prediction tasks. Although some studies display a slight superiority of GRU architectures versus LSTM networks, (around 1% increase in AUC), other studies did not find significant differences between them. Overall, LSTM and GRU have similar memory-retention mechanisms, though GRU implementations are less complex and have faster runtimes [89]. Due to this similarity, most works have used one without benchmarking it against the other. In addition, for very long EHR sequences, such as ICU admissions with a high rate of recorded medical events, bidirectional GRU and LSTM networks consistently outperformed their unidirectional counterparts. This is likely due to the fact that bidirectional recurrent networks simultaneously learn from both past and future values in a temporal sequence, and so, they retain additional trend information [69]. This is particularly important in the healthcare context, since patient health status patterns change rapidly or gradually over time [12]. For example, an ICU patient with a rapidly fluctuating health status over the past week may eventually die, even if the patient is currently in a good condition. Another patient, initially admitted to the ICU within the past week in a very bad condition, may gradually improve and survive. Therefore, bidirectional recurrent networks are the most state-of-the-art DL models for time series prediction in healthcare. GRU, which has lower complexity and comparable performance to LSTM, is the preferred model variant, though additional comparative studies are recommended by this review to affirm this conclusion.

While a majority of the RNN studies employed single-layer architectures, some studies chose an increased complexity with multi-layered GRU [7,48], LSTM [40,64,68,74], and Bi-GRU [2,67,82] networks. Other than two earlier works [7,48], multi-layered architectures were not consistently tested against their single-layered counterparts. Consequently, it is difficult to decipher if adding additional RNN layers, whether they are bidirectional or not, improves learning performance. On the other hand, *channel-wise learning*, a technique which trains a separate RNN layer per feature or feature type, successfully enhanced traditional RNN models which contain network layers that learn all feature parameters simultaneously. There are two underlying ideas behind this development. First, it helps identify unique patterns within each individual time-series (e.g. body organ system status) [17], prior to integration with patterns found in multivariate data. Second, channel-wise learning facilitates the identification of patterns related to *informed missingness*, by discovering which of the masked variables correlates strongly with other variables, target or otherwise [12]. Nevertheless, channel-wise needs further benchmarking against vanilla RNN models to learn the conditions under which it is most beneficial. Additionally, certain works enhanced upon the supervised learning process of RNN models. For prediction tasks with a static target, such as in-hospital mortality, RNN models were supervised at multiple time steps instead of merely the final time-point. This so-called *target replication* has been shown to be quite efficient during backpropagation [64]. Specifically, instead of passing patient target information across many time-steps, the prediction targets are replicated at each time-point within the sequence, thus providing additional local error signals which can be individually optimized. Moreover, target replication can improve model predictions even when the temporal sequence is perturbed by small, yet significant, truncations.

As noted in Section 3.3, convolutional network models were more commonly used in the early stages of deep time series prediction for healthcare. Eventually, they were shown to be consistently outperformed by recurrent models. However, recent architectural trends have been using convolutional layers as a complement to GRU and LSTM [44,54,59,73]. The underlying idea is that RNN layers capture the global structure of the data via modeling interactions between events, while the CNN layers, using their temporal convolution operators [54], capture local structures of the data occurring at various abstraction levels. Therefore, our systematic review suggests using CNNs to enhance RNN prediction performance, instead of employing either in a standalone setting. Another recent trend in the literature is the splitting of entire temporal sequences into subsequences for various time periods—before applying convolutions of different filter size—in order to capture temporal patterns within each time period [49]. For optimal local pattern (motif) detection, slow-fusion CNN which considers both individual patterns of the time-periods as well as their interactions has been shown to be the most effective convolutional approach [18].

Several important research gaps were identified in the models used for deep time series prediction in healthcare. First, there is no systematic comparison among state-of-the-art models in different healthcare settings, such as rare versus common diseases, chronic versus non-chronic maladies, and inpatient versus outpatient visits. These different healthcare settings have identifiable heterogeneous temporal data characteristics. For instance, outpatient EHR data contains large numbers of visits with few medical events recorded during each visit, while inpatient visits contain relatively few visit records but documented long sequences of events for each visit. Therefore, the effectiveness of a given DL architecture will vary over these different clinical settings. Second, it is not clear whether adding multiple layers of RNN or CNN within a given architecture can further improve model performance. The maximum number of layers observed within the reviewed and selected studies was two. Given enough training samples, the addition of more layers may further improve performance by allowing for the learning of increasingly sophisticated temporal patterns. Third, a majority of the reviewed studies (92%) targeted a prediction task on EHR data, while the generalizability of the models to AC data needs more investigation. For example, although many studies reported promising outcomes for EHR-based hospital readmission predictions using GRU models, Min et al. (2019)[1] found that similar DL architectures are ineffective for claims data. Finding novel models which can extract temporal patterns from EHR data—which are simultaneously applicable to claims data—can be an interesting future direction for transfer learning projects. Fourth, while channel-wise learning seems to be a promising new trend, it needs researchers to further investigate the precise temporal patterns detected by this approach. DL methods focused on interpretability would be ideal for such an application. Fifth, many studies compared their DL methods against expert domain knowledge, but a hybrid approach that leverages expert domain knowledge within the embeddings should help improve representation performance. Lastly, the prediction of medications, either by code or group, has been a well-targeted task. However, a more aggressive approach, such as predicting medications along with their appropriate dosage and frequency, would be a more realistic and useful target for clinical decision making in practice.

### 4.4 Addressing temporal irregularity

The most common approach for handling visit irregularity is to treat the time interval between adjacent events as an independent variable and concatenating it to the input embedding vectors. While this technique is easy to implement, it does not consider contextual differences between recent and earlier visits. Addressing this limitation, researchers modified the internal memory cells of RNN networks by giving higher weights to recent visits [20,65]. However, a systematic comparison between the two approaches has not been explored. Therefore the time interval approach, which has been shown to be effective in various applications, remains the most efficient, tested strategy to handle visit irregularity. It is noteworthy that tokenizing time intervals is also considered the most effective method of capturing duration in the context of natural language processing [110,111], a field of study which inspires many of the deep time series prediction methods in healthcare.

Although most works addressing irregularity focus on visit irregularity, a few studies concentrated on feature irregularity [60,91]. A fundamental concept underpinning the difference between the two is that fine-grained temporal information is more complex and yet important to learn at the feature level than visit level. Specifically, different features expose different temporal patterns, such as when certain features decay faster than others. Paralleling the work on visit irregularity and time intervals, these studies [60,91] modified the internal processes of RNN networks to learn unique decay patterns for each individual input feature. Again, this research direction is relatively new and boasts few published works, so it is difficult to make a general suggestion for unilaterally handling feature irregularity in deep time series learning tasks.

Overall, we can stipulate that adjusting the memory mechanisms of recurrent networks when addressing either visit or feature irregularity needs additional benchmarking experiments in order to make the arguments robust. Currently, it has been evaluated in a single hospital setting for each case. Therefore, optimal synergies among patient types (inpatient vs. outpatient), sequence lengths (long vs. short) and irregularity approaches (time interval vs. modifying RNN memory cells) are not entirely conclusive, but time interval approaches have been most commonly published

### 4.5 Attention mechanisms

Attention mechanisms have been employed by researchers with the premise that neither patient visits nor medical codes should contribute equally when performing a target prediction task. As such, learning attention weights for visits and codes have been the subject of many deep time series prediction studies. The three most commonly used attention mechanisms are (1) location-based, (2) general attention, and (3) concatenation-based frameworks. The methods differ primarily on how the learned weight parameters are connected to the model's hidden states [69]. Location-based attention schemes calculate weights from the most current hidden state. Alternatively, general attention calculations are based on a linear combination connecting the current hidden states to the previous hidden states, with weight parameters being the linear coefficients. Most complex is the concatenation-based attention framework, which trains a multilayer perceptron to learn the relationship between parameter weights and hidden states. Location-based attention systems have been the most commonly used attention mechanisms for deep time series prediction in healthcare.

We found several research gaps regarding attention. Most studies relied on attention mechanisms to improve the interpretability of their proposed DL model by highlighting important visits or medical codes, without evaluating the differential effect of attention on prediction performance. This is an important issue, as incorporating attention into a model may improve interpretability, but it does not have an established effect on performance in the DL for healthcare time series domain. Furthermore, with only a single exception [57], we did not find studies reporting the separate contributions of visit-level attention and medical code-level attention. Lastly, and again with only a single exception [69], no study compared the performance or interpretability of different attention mechanisms. All of these research gaps should be investigated in a comprehensive manner in future studies, particularly for EHR data, as most prior attention studies have focused on the clinical histories of individual patients.

### 4.6 Incorporation of medical ontologies

When incorporating medical domain knowledge into deep time series prediction models, researchers have mainly utilized medical ontology trees and knowledge graphs within embedding layers of recurrent networks. Some of the success of these approaches is due to the enhancement they provide when addressing rare diseases. Being less frequent in the data, a proper representation and pattern extraction for rare diseases is challenging for simple RNN models. Medical domain knowledge graphs provide rare disease information to the model through ancestral node embeddings that contain hierarchical information of the disease. However, this advantage is not as exceptional when sufficient data is available for all patients over a long record history [53,70]. Continuing research is needed to expand the innovative architectures that incorporate medical ontologies for a broad variety of prediction tasks and case studies.

### 4.7 Static data inclusion

There are four published approaches for integrating static patient data with their temporal data. By far, the most common approach is to feed a vector of static features as additional input to the final fully connected layer of a DL network. Another strategy trains a separate feedforward neural network on the static features, and then adds the encoded output of this separate network to the final dense layer in the principal neural network for target prediction. Researchers have also injected static data vectors as input to each time point of the recurrent network, effectively treating the patient demographic and historical data as quasi-dynamic. Lastly, similar to those strategies that handle visit and feature irregularities, researchers have modified the internal memory processes of recurrent networks to incorporate specific static features as input.

The most important research gap regarding static data inclusion is that we have found no study evaluating the differential effects of static data on prediction performance. Moreover, comparing these four approaches in a meaningful benchmarking setting, with the expressed goal of finding the most optimal technique, could be an interesting future research direction. Finally, since DL models may not learn the same representation for every subpopulation of patients (e.g., male vs. female, chronic vs. non-chronic, or young vs. old), significant research gaps exist in the post-analysis of static feature performance as input. Such analyses could inform decision makers of crucial insights into model fairness, and would also stimulate future research on predictive models that better balances fairness with accuracy.

### 4.8 Learning strategies

Recent literature has investigated three new DL strategies: (1) cost-sensitive learning, (2) multi-task learning, and (3) transfer learning. While many reviewed studies used an imbalanced dataset for their experiments, a select few embedded cost information as a learning strategy that incorporated additional cost-sensitive loss. Specifically, each of these studies changed the loss function of the DL model to increasingly penalize for misclassification of the minority class. In the healthcare domain, imbalanced datasets are very common, and patients with diseases are less common than healthy patients. Moreover, most of the prediction tasks on the minority class lead to critical care decisions, such as identifying those patients who are likely to die in the next 48 hours or those who will become diabetic in the relatively near future. Devising cost-sensitive learning components into DL networks thus needs further attention and is a wide-open research gap for future inquiry. As an example, exploring cost-sensitive methods in tandem with the traditional ML techniques of oversampling or underdamping could lead to significant performance increases in model prediction rates for the minority class. In addition, calculating the precise cost savings when correctly identifying the minority class of patients, similar to [61], can further underline the importance of the cost-sensitive learning strategy.

Researchers have reported the benefit of multi-task learning by documenting its performance in a significant variety of healthcare outcome prediction tasks. However, the cited works do not distinguish the model components that exemplify why learning a single, multi-task deep model is preferable to simultaneously training multiple DL models for respective individualized prediction tasks. More specifically, we ask which layers, components, or learned temporal patterns in a DL network should be shared among different tasks, and in which healthcare applications might this strategy be most efficient? These research questions are straightforward and could be fruitfully studied in the near future with explainable DL models.

Among the three noted, transfer learning was the least studied learning strategy found within our systematic review of the literature with just a single citation [43], displaying the effectiveness of the method for both task and domain adaptation. It is commonly assumed that, with sufficient data, trained DL models can be effective for a wider variety of prediction tasks and domains. However, in many healthcare settings, such as those with rural patients, sufficient data is difficult to collect [112]. Transfer learning methods have the potential to make a huge impact on deep time series prediction in healthcare by making pre-trained models applicable to essentially any healthcare setting. Still, further research is recommended to ascertain which pathological prediction tasks are most transferable, which network architectures are most flexible, and which model parameters require the least tuning when transferring to different domains.

### 4.9 Interpretations

One of the most common critiques of DL models is the difficulty of their interpretation, and researchers have attempted to alleviate this issue with five different approaches. The first approach uses feature importance measures such as Shapley and DeepLIFT. A Shapley value of a feature is the average of its contribution across all possible coalitions with other features, while DeepLIFT compares the activation of each neuron in the deep model inputs to its default reference activation value and assigns contribution scores according to the difference [113]. Although both of these measures cannot illuminate the internal procedure of DL models, they can identify which features have been most frequently used to make final predictions. A second approach visualizes what input data the model focused on for each individual patient [13] through the implementation of interpretable attention mechanisms. Particularly, some studies investigated which medical visits and features contributed most to prediction performance with a network attention layer. As a clinical decision support tool, this raises clinician awareness of which medical visits deserve careful human examination. In addition to individual patient visualization, a third interpretation tactic aggregated model attention weights to calculate the most important medical features for specific diseases or patient groups. Additionally, error analysis of final prediction results allowed for consideration of the medical conditions or patient groups for which a DL model might be more accurate. This fourth interpretation approach is also popular in non-healthcare domains [114]. Finally, considering each set of medical events as a basket of items and each target disease as the label, researchers extracted frequent patterns of medical events most predictive of the target disease.

Overall, this review found explainable attention to be the most commonly used strategy for interpreting deep time series prediction models evaluated on healthcare applications. Indeed, individual patient exploration can help make DL models more trustworthy to clinicians and facilitate subsequent clinical actions. Nevertheless, because implementing feature importance measures is much less complex, this study recommends consistently reporting them on most healthcare deep times series prediction studies, providing useful clinical implication with little added effort. Although individual-level interpretation is important, extracting general patterns and medical events associated with target healthcare outcomes is also beneficial for clinical decision makers, thereby contributing to clinical practice guidelines. We found just one study implementing a population-level interpretation [63], extracting frequent CNN motifs of medical codes associated with different diseases. Otherwise, researchers broadly have reported the top medical codes with the highest attention weights for all patients [2] or different patient groups, in order to provide a population-level interpretation. This current limitation can be an essential direction for future research involving network interpretability.

### 4.10 Scalability

We identified two main findings regarding the scalability of deep time series prediction methods in healthcare. First, although DL models are usually evaluated on a single dataset with a limited number of features, some studies confirmed their scalability to large hospital EHR datasets with high dimensionality. The fundamental observation is that higher dimensionality and larger amounts of data can further enhance model performance by raising their representational learning power [42]. Such studies have typically used single-layered GRU or LSTM architectures, but analyzing more advanced neural network schemas, such as those proposed in recent studies (Section 3.1) is a venue for future research. Also, one scalability study observed that models which are primarily purposed for EHR data may not be as effective with AC data [1]. This is mainly because potent predictive features available in EHR data, such as lab test results, tend to be missing in AC datasets. Therefore, scalability studies on AC data merits further inquiry. Second, DL models are typically compared against traditional supervised ML methods on a singular method only (Table S3). However, two studies [1,42] compared DL methods against ensembled traditional supervised learning models, both on EHR and AC data, and found their performances are comparable. This shows an important research gap for proper comparison between DL and traditional supervised learning models to identify data settings, such as feature types, dimensionality and missingness, in which DL models either perform comparably or excel against their traditional ML counterparts.

## 5 CONCLUSION

In this work, we systematically reviewed studies focused on deep time series prediction to leverage structured patient time series data for healthcare prediction tasks from a technical perspective. The following is a summary of our main findings and suggestions:

- *Patient representation*: There are two common approaches: sequence representation and matrix representation. For prediction tasks in which inputs are numeric, such as lab tests or vital signs, sequence representations have typically been used. For those with categorical inputs, such as diagnosis codes or procedure codes, matrix representation is the premiere choice. In order to combine numeric and categorical inputs, researchers have employed three distinct methods: (1) assigning a unique token to each combination of measure name, value, and unit; (2) encoding the numeric measures categorically as missing, low, normal, or high; and (3) converting the numeric measures to severity scores to further discretize them as low, normal, or high. Moreover, embedding medical events in a sequence representation involved an additional three prevailing techniques: (1) adding a separate embedding layer to learn an optimal medical code representation from scratch, (2) adopting a pre-trained embedding layer such as with word2vec, or (3) using a medical code grouping strategy, sometimes involving clinical classification software. Comparing these diverse approaches and techniques in a solid benchmarking setting needs further investigation.

- *Missing value handling*: Missing values in healthcare data are generally not missing at random, but often reflect decisions by caregivers. Capturing missing values as a separate input masking vector or learning the missing patterns with a neural network have been the most effective methods to date. Identifying impactful missing features will help healthcare providers implement optimal data collection strategies and better inform clinical decision-making.
- *Deep learning models*: RNN architectures, particularly their single-layered GRU and LSTM versions, were identified as the most prevalent networks in extant literature. These models excel at large sequences of input data representing longitudinal patient history. While RNN models extract global temporal patterns, CNNs are proficient at detecting local patterns and motifs. Combining RNN and CNN in a hybrid structure for capturing both types of patterns has become a trend in recent studies. More investigation is required to understand optimal network architecture for various hospital settings and learning tasks.
- *Addressing temporal irregularity*: For handling visit irregularity, the time interval between visits is given as an additional independent input, or alternatively, the internal memory processes of recurrent networks are slightly modified to assign differing weights to earlier versus more recent visits. When addressing feature irregularities, the memory and gating activities of RNN networks are similarly modified to learn individualized decay patterns for each feature or feature type. Overall, temporal irregularity handling methods need more robust benchmarking experiments in an assortment of hospital settings, including variations in patient type (inpatient vs. outpatient) and visit length (long-sequence vs. short-sequence).
- *Attention mechanisms*: Location-based attention is by far the most commonly used means of differentiating importance in portions of the input data and network nodes. Most studies used attention mechanisms to improve the interpretability of their proposed DL models by highlighting important visits or medical codes, but without evaluating the differential effect of attention mechanisms on prediction performance. Furthermore, we found that further inquiry is warranted to separately evaluate contributions of visit-level and medical code attention.
- *Incorporation of medical ontologies*: Researchers have incorporated medical ontology trees and knowledge graphs in the embedding layers of recurrent networks to compensate for lack of sufficient data when regarding rare diseases for prediction tasks. Using these medical domain knowledge resources, the information for such rare diseases is captured through the ancestral nodes and pathways in the tree or graph for input into network embeddings.
- *Static data inclusion*: We found four basic approaches followed by researchers to merge demographic and patient history data with the dynamic longitudinal input of EHR or AC data: (1) feeding static features to the final fully-connected layer of the neural network, (2) training a separate feedforward network for the subsequent inclusion of encoded output into the main network, (3) the repetition of static feature input at each time point in a quasi-static manner, and (4) modifying the internal processes of recurrent networks. We found no study evaluating the effects of static data on prediction performance, especially post analysis of performance results for static features.
- *Learning strategy*: Three learning strategies have been investigated by the authors included in this review: (1) cost-sensitive, (2) multi-task, and (3) transfer learning. Devising cost-sensitive learning components into DL networks is a wide-open research gap for future study. Regarding multi-task learning, researchers have reported its benefit by citing increased performance levels in a variety of healthcare outcome prediction tasks. However, multi-task learning does not make clear which network layers, components, or types of extracted temporal patterns within the architectural design should be shared among the different tasks—as well as in which healthcare scenarios the multi-task strategy is most efficient. Transfer learning was the least studied method found in our systematic review, but it has promising prospects for further inquiry, as the scale of data and number of external data sets in published works increases.

- *Interpretations*: The most common approach to visualize important visits or medical codes on individual patients was the use of an attention mechanism in the neural network. Although individual-level interpretation is indeed important, as a future research direction, the use of population-level interpretation techniques to extract general patterns and identify specific medical events associated with target healthcare outcomes will be a boon for clinical decision makers.
- *Scalability*: Several studies confirm the generalizability of well-known deep time series prediction models to large hospital EHR datasets, even with high input dimensionality. However, analyzing advanced network architectures that have been proposed in recent works is a suggested venue for future research. Furthermore, some studies found that ensembles of traditional supervised learning methods have comparable performance to DL models, both on EHR and AC data. Important research gaps remain for establishing a proper comparison of DL against single or ensembled traditional ML models. In particular, it would be useful to identify patient, dimensionality, and missing value conditions in which DL models, with their higher complexity and runtimes, might be superfluous. This is a continual concern when considering the need for implementing real-time information systems that can better inform clinical decision makers.

A potential limitation of this systematic review is a possible incomplete retrieval of relevant studies on deep time series prediction. Although we included a wide set of keywords, it remains challenging to conduct an inclusive search strategy with an automatic query by keyword searching. We alleviated this concern by applying snowballing search strategies from the originally included publications. In other words, we assumed that any newer publication should reference one of the former included studies within their paper, especially well-known benchmarking models such as Doctor AI [7], RETAIN [22], DeepCare [65], and Deepr [63]. Another challenge was selectively differentiating the included studies from numerous other adjacent works when predicting a single clinical outcome with a DL methodology. To achieve this, we implemented a full-text review step that included all papers which specifically mention patient representations or embedding strategies. Additionally we ensured that the authors' stated goals involved learning these representations at a patient level, and not merely devising models to maximize performance on a specific disease prediction task. The aforementioned limitations pose a potential threat to selective bias in publication trends for any systematic review, but particularly one in which publication rates increase with recency, such as seen in the ever-increasing popularity of utilizing DL models on a myriad of applications, healthcare or otherwise.

Table S1: Search strategy.

| Database | Search query |
|---|---|
| ACM digital library | (”electronic health record” OR ”electronic health records” OR ”EHR” OR ”EHRs” OR ”electronic medical record” OR ”electronic medical records” OR ”EMR” OR ”EMRs” OR ”medical claim” OR ”medical claims” OR ”medical record” OR ”medical records” OR ”administrative claims” OR ”administrative claim” OR ”administrative data” OR ”administrative health data” OR ”administrative healthcare data” OR ”administrative health care data” OR ”pharmacy claim” OR ”pharmacy claims” OR ”claims data” OR ”claim data”) AND (”Convolutional Neural Networks” OR ”Convolutional Neural Network” OR ”CNN” OR ”CNNs” OR ”Recurrent Neural Networks” OR ”Recurrent Neural Network” OR RNN* OR ”long short-term memory” OR LSTM* OR ”bidirectional LSTM” OR ”bidirectional LSTMs” OR ”Bi-LSTM” OR ”Bi-LSTMs” OR ”gated recurrent unit” OR ”GRU” OR ”GRUs” OR “deep learning”) |
| IEEE | (("Full Text & Metadata": "electronic health record") OR ("Full Text & Metadata”: “electronic health records") OR ("Full Text & Metadata”: “EHR") OR ("Full Text & Metadata”: “EHRs") OR ("Full Text & Metadata”: “electronic medical record") OR ("Full Text & Metadata”: “electronic medical records") OR ("Full Text & Metadata”: “EMR") OR ("Full Text & Metadata”: “EMRs") OR ("Full Text & Metadata”: “medical claim") OR ("Full Text & Metadata”: “medical claims") OR ("Full Text & Metadata”: “medical record") OR ("Full Text & Metadata”: “medical records") OR ("Full Text & Metadata”: “administrative claims") OR ("Full Text & Metadata”: “administrative claim") OR ("Full Text & Metadata”: “administrative data") OR ("Full Text & Metadata”: “administrative health data") OR ("Full Text & Metadata”: “administrative healthcare data") OR ("Full Text & Metadata”: “administrative health care data") OR ("Full Text & Metadata”: “pharmacy claim") OR ("Full Text & Metadata”: “pharmacy claims") ) AND (("Full Text & Metadata”: “Convolutional Neural Networks") OR ("Full Text & Metadata”: “Convolutional Neural Network") OR ("Full Text & Metadata”: “CNN") OR ("Full Text & Metadata”: “CNNs") OR ("Full Text & Metadata”: “Recurrent Neural Networks") OR ("Full Text & Metadata”: “Recurrent Neural Network") OR ("Full Text & Metadata”: “RNN") OR ("Full Text & Metadata”: “RNNs") OR ("Full Text & Metadata”: “long short-term memory") OR ("Full Text & Metadata”: “LSTM") OR ("Full Text & Metadata”: “LSTMs") OR ("Full Text & Metadata”: “bidirectional LSTM") OR ("Full Text & Metadata”: “bidirectional LSTMs") OR ("Full Text & Metadata”: “Bi-LSTM") OR ("Full Text & Metadata”: “Bi-LSTMs") OR ("Full Text & Metadata”: “gated recurrent unit") OR ("Full Text & Metadata”: “GRU") OR ("Full Text & Metadata”: “GRUs") OR ("Full Text & Metadata”: “deep learning")) |
| Scopus | (”electronic health record” OR ”electronic health records” OR EHR* OR ”electronic medical record” OR ”electronic medical records” OR EMR* OR ”medical claim” OR ”medical claims” OR ”medical record” OR ”medical records” OR ”administrative claims” OR ”administrative claim” OR ”administrative data” OR ”administrative health data” OR ”administrative healthcare data” OR ”administrative health care data” OR ”pharmacy claim” OR ”pharmacy claims” OR ”claims data” OR ”claim data”) AND (”Convolutional Neural Networks” OR ”Convolutional Neural Network” OR CNN* OR ”Recurrent Neural Networks” OR ”Recurrent Neural Network” OR RNN* OR ”long short-term memory” OR LSTM* OR ”bidirectional LSTM” OR ”bidirectional LSTMs” OR Bi-LSTM* OR ”gated recurrent unit” OR GRU* OR “deep learning”) |
| MEDLINE | (("electronic health record”[All Fields]) OR ("electronic health records”[All Fields]) OR ("EHR”[All Fields]) OR ("EHRs”[All Fields]) OR ("electronic medical record”[All Fields]) OR ("electronic medical records”[All Fields]) OR ("EMR”[All Fields]) OR ("EMRs”[All Fields]) OR ("medical claim”[All Fields]) OR ("medical claims”[All Fields]) OR ("medical record”[All Fields]) OR ("medical records”[All Fields]) OR ("administrative claims”[All Fields]) OR ("administrative claim”[All Fields]) OR ("administrative data”[All Fields]) OR ("administrative health data”[All Fields]) OR ("administrative healthcare data”[All Fields]) OR ("administrative health care data”[All Fields]) OR ("pharmacy claim”[All Fields]) OR ("pharmacy claims”[All Fields] OR ("claims data”[All Fields] OR ("claim data”[All Fields]) OR ("claim data”[All Fields]) OR ("claims data”[All Fields])) AND (("Convolutional Neural Networks”[All Fields]) OR ("Convolutional Neural Network”[All Fields]) OR ("CNN”[All Fields]) OR ("CNNs”[All Fields]) OR ("Recurrent Neural Networks”[All Fields]) OR ("Recurrent Neural Network”[All Fields]) OR ("RNN”[All Fields]) OR ("RNNs”[All Fields]) OR ("long short-term memory”[All Fields]) OR ("LSTM”[All Fields]) OR ("LSTMs”[All Fields]) OR ("bidirectional LSTM”[All Fields]) OR ("bidirectional LSTMs”[All Fields]) OR ("Bi-LSTM”[All Fields]) OR ("Bi-LSTMs”[All Fields]) OR ("gated recurrent unit”[All Fields]) OR ("GRU”[All Fields]) OR ("GRUs”[All Fields]) OR ("deep learning”[All Fields]) ) |

Table S2: List of abbreviations.

| Abbreviation | Full | Abbreviation | Full |
|---|---|---|---|
| VS | Vital Sign | RF | Random Forest |
| LT | Lab Test | LDA | Linear Discriminant Analysis |
| Dm | Demographic | KNN | K-Nearest Neighbor |
| Px | Procedure Code | SVM | Support Vector Machine |
| Dx | Diagnosis Code | LR | Linear Regression |
| Rx | Medication Code | MLP | Multi-Layer Perceptron |
| EHR | Electronic Health Record | NB | Naïve Bayesian |
| AC | Administrative Claims | RF | Random Forest |
| | | DT | Decision Tree |

Table S3: Reviewed studies' extracted features according to Table 1.

| Author | Medical Task | Database | Data Size | Data Type | VS | LT | Dm | Px | Dx | Rx | Others | Patient Representation | Windowing | Missing value | Architecture | Output Temporality | Performance | Benchmark | Interpretation |
|---|---|---|---|---|---|---|---|---|---|---|---|---|---|---|---|---|---|---|---|
| Zhang et al. (2018)[57] | Hospital re-admissions in 30 days prediction | University of Virginia Health System | 97,942 patients | EHR | - | - | Y | Y | Y | Y | - | Multi-hot vector | Every 90 days aggregated by a convolution layer | - | 1: Input<br>2: Embedding with Skip-Gram<br>3: Convolution<br>4: Bi-GRU with attention<br>5: Dense | temporal | AUC=0.799 | LR, MLP, GRU, RETAIN | Temporal visualization of a patient visit records generated by attention mechanism + Population level feature importance based on attention mechanism |
| Cheng et al. (2016)[18] | Congestive heart failure and chronic obstructive pulmonary in six month prediction | Sutter Palo Alto Medical Foundation Primary Care | 7,839 patients | EHR | - | - | - | - | Y | - | - | Temporal matrix | Daily mean | - | 1: Input<br>2: Convolution<br>3: Dense | temporal | AUC=0.767<br>AUC=0.738 | LR | Aggregating weights of the neurons assigned to each medical feature to obtain the importance |
| Lipton et al. (2015)[64] | Diagnoses prediction | Children's Hospital in LA | 10,401 ICU admissions | EHR | Y | Y | - | - | - | - | - | Temporal matrix | Hourly mean | Forward imputation | 1: Input<br>2: LSTM with target replication<br>3: LSTM with target replication<br>4: Dense | temporal | AUC=0.856 | LR, MLP | - |
| Choi et al. (2016)[7] | Diagnosis, medication and time to next visit prediction | Sutter Health Palo Alto Medical Foundation Primary Care | 263,706 patients | EHR | - | - | - | Y | Y | Y | Time since last visit | Multi-hot vector | - | - | Doctor AI:<br><br>1: Input<br>2: Embedding with Skip-Gram<br>3: GRU<br>4: GRU<br>5: Dense | temporal | Recall@30=0.795<br>Recall@30=0.797<br>R2=0.253 | LR, MLP | - |
| Pham et al. (2016)[65] | Diagnosis prediction for mental health and diabetes | An Australian Hospital | 13,300 patients | EHR | - | - | - | Y | Y | Y | Time since last visit and admission type | Multi-hot vector on code groups | - | - | DeepCare :<br><br>1: Input<br>2: Embedding<br>3: Modified LSTM with decay and admission | temporal | Precision@1=52.7<br>Precision@1=66.2 | SVM, RF, RNN, LSTM | - |
| | | | | | | | | | | | | | | | type with attention mechanism<br>4: Dense | | | | |
| Nguyen et al. (2017)[63] | Unplanned readmission in 6 months prediction | A Private Hospital in Australia | 9,986 patients | EHR | - | - | - | Y | Y | - | Discretized time since ast visit | Multi-hot vector | - | - | Deepr :<br>1: Input<br>2: Embedding with Word2Vec<br>3: CNN<br>4: Dense | temporal | AUC=0.809 | LR | Example list of frequent patterns for different disease |
| Esteban et al. (2016)[3] | Kidney transplantation endpoint in 6 months prediction | Charite Hospital in Berlin | 2,061 patients | EHR | - | Y | Y | - | - | Y | 342 static features | Multi-hot vector | - | Zero value imputation | 1: Input<br>2: Embedding<br>3: GRU<br>4: MLP with patients' static data<br>5: Dense | temporal | AUC=0.833 | LR, LSTM | - |
| Che et al. (2018)[36] | Mortality, Length of stay, surgery recovery, cardiac condition prediction | MIMIC-II | 8,000 admissions | EHR | Y | Y | - | - | - | Y | - | Temporal matrix | Hourly mean | Learned function of the average and the latest value | 1: Input<br>2: Modified GRU with decay<br>3: Dense | static and temporal | AUC= 0.837 (Avg of 4 tasks) | LR, MLP, , RF, SVM, KNN, LSTM, GRU, Doctor AI, DeepCare | - |
| Choi et al. (2017)[46] | Heart failure in 6 months prediction | Sutter Health Palo Alto Medical Foundation Primary Care | 333,17 patients | EHR | - | - | - | Y | Y | Y | Time since last visit | Multi-hot vector | - | - | 1: Input<br>2: Embedding with Skip-Gram<br>3: Dense<br>4: Logistic regression | temporal | AUC=0.777 | LR, MLP, SVM, KNN | - |
| Sun et al. (2019)[9] | Mortality prediction | MIMIC-III and eICU | 33,798 and 98,897 patients | EHR | Y | Y | - | - | - | - | - | Temporal matrix | Hourly mean | Mean and latest value imputation | 1: Input<br>2: Bi-GRU<br>3: Dense | static | AUC=0.860<br>AUC= 0.845 | LR, GRU | - |
| Wang et al. (2019)[20] | Diagnoses prediction | MIMIC-III | 7,537 patients | EHR | - | - | - | - | Y | - | Time since last visit | Multi-hot vector | - | - | 1: Input<br>2: Embedding with enhanced Skip-Gram<br>3: Modified LSTM with time interval<br>4: Dense | temporal | Recall@20 = 0.782 | LSTM | - |
| Yang et al. (2019)[66] | Diagnoses prediction | MIMIC-III | 33,798 patients | EHR | Y | Y | - | - | - | - | - | Multi-hot vector | Hourly mean | Forward imputation | 1: Input<br>2: Embedding<br>3: Bi-LSTM<br>4: Dense | temporal | Precision=0.867<br>Recall=0.801 | LR, MLP | Temporal visualization of a patient visit records generated by |

| | | | | | | | | | | | | | | | | | | |
|---|---|---|---|---|---|---|---|---|---|---|---|---|---|---|---|---|---|---|
| | | | | | | | | | | | | | | | | | | attention mechanism |
| Wang et al. (2018)[58] | Readmission in 30 and 60 days prediction | Washington University School of Medicine and Barnes-Jewish Hospital | 2,565 patient | EHR | Y | Y | Y | - | - | - | - | Temporal matrix | First and second order features | Median value imputation | 1: Input<br>2: Embedding<br>3: CNN<br>4: CNN<br>5: Dense | temporal | AUC=0.440<br>AUC=0.450 | - | - |
| Chen et al. (2018)[17] | Illness severity prediction | MIMIC-III | 36,740 admissions | EHR | Y | Y | - | - | - | - | - | Temporal matrix | Hourly mean | Forward imputation | 1: Input<br>2 to k+1: *k* GRUs for each body organ<br>k+2: Dense | static | Precision =0.888<br>Recall =0.897 | SVM, RF, DT, LDA, XGBoost | - |
| Park et al. (2018)[82] | Vascular diseases prediction | Seoul National University Bundang Hospital | 45,315 patients | EHR | - | - | - | - | Y | Y | Visit type and Time since last visit | Multi-hot vector | - | - | 1: Input<br>2: Embedding<br>3: Bi-GRU with attention mechanism for medications<br>4: Bi-GRU with attention mechanism for diagnosis<br>5: Dense | static | AUC=0.848 | RETAIN | Grad-CAM for an individual patient |
| Guo et al. (2019)[67] | Diagnoses prediction | A Hospital in China and MIMIC-III | 46,520 and 43,853 patients | EHR | - | - | - | - | Y | Y | - | Multi-hot vector | - | - | 1: Input<br>2: Embedding for diagnosis<br>3: Embedding for medications<br>4: Bi-GRU for diagnosis<br>5: Bi-GRU for medications<br>6: Crossover attention<br>7: Dense | temporal | F-1=0.666<br>F-1=0.326 | GRU, RETAIN, Dipole | Temporal visualization of a patient visit records generated by attention mechanism |
| Yin et al. (2019)[47] | Heart failure prediction | MIMIC-III | 1,700 patients | EHR | - | - | - | - | Y | - | Time since last visit and time to the index date | Multi-hot vector | - | - | 1: Input<br>2: Embedding for diagnosis and time<br>3: LSTM for diagnosis and time | temporal | AUC=0.737 | LR, RF, SVM, GRU, LSTM, REATIN, GRAM, KAME | Temporal visualization of a patient visit records generated by attention mechanism + Population |
| | | | | | | | | | | | | | | | 4: LSTM for time and diagnosis with knowledge attention mechanism<br>5: Dense | | | | level feature importance based on average attention |
| Wang et al. (2018)[48] | Hearth failure in 3 months prediction | A Hospital in China | 28,753 patients | EHR | - | - | Y | - | Y | Y | - | Multi-hot vector | - | - | 1: Input<br>2: GRU<br>3: GRU<br>4: Dense | temporal | F-1=0.767 | LR, MLP, RNN | - |
| Ju et al. (2020)[49] | Heart failure and diabetes in 3 months prediction | IBM T. J. Watson Research Center | 15,120 patients | EHR | - | - | Y | - | Y | Y | Smoking status | Multi-hot vector | - | - | 1: Input<br>2: Embedding with Word2Vec<br>3: CNN<br>4: CNN<br>5: Dense | temporal | AUC=0.943<br>AUC=0.999 | LR, RF, SVM, GRU, LSTM, CNN | - |
| Wang et al. (2019)[68] | Diagnoses prediction | MIMIC-III | 7,105 patients | EHR | - | - | Y | - | Y | - | - | Multi-hot vector | - | - | 1: Input<br>2: LSTM<br>3: LSTM<br>3: Dense | temporal | F-1=0.991 | GRU | - |
| Gao et al. (2019)[100] | Neonatal Encephalopathy prediction | Vanderbilt University Medical Center | 31,158 patients | EHR | - | - | - | - | Y | - | - | Multi-hot vector | - | - | 1: Input<br>2: Embedding with Skip-Gram<br>3: LSTM<br>4: Dense | static | AUC=0.933 | LR | - |
| Zhang et al. (2017)[83] | Septic shock prediction | Christiana Care Health System | 12,980 visits | EHR | Y | Y | - | - | - | - | Location data | Temporal matrix | - | Forward imputation | 1: Input<br>2: LSTM<br>3: Dense | temporal | F-1=0.961 | LR, SVM, RNN, GRU | - |
| Yu et al. (2020)[37] | Mortality prediction | Stanford Translational Research Integrated Database Environment | 32,604 patients | EHR | Y | Y | Y | - | - | - | - | Temporal matrix | Hourly mean | Forward imputation | 1: Input<br>2: LSTM with attention mechanism<br>3: Dense | static | AUPRC=0.520 | - | - |
| Lin et al. (2019)[59] | Readmission in 30 days prediction | MIMIC-III | 35,334 patients | EHR | Y | Y | Y | - | Y | - | - | Multi-hot vector | Hourly mean | Forward imputation | 1: Input<br>2: Embedding<br>3: Bi-LSTM<br>4: CNN<br>5: Dense | temporal | AUC=0.548 | LR, NB, RF, SVM, LSTM, CNN | - |
| Duan et al. (2019)[81] | Main adverse cardiac events prediction | Cardiology Department of a Chinese | 2,930 patients | EHR | - | Y | Y | - | Y | - | - | Multi-hot vector | - | - | 1: Input<br>2: Embedding<br>3: Bi-LSTM | static | Accuracy=0.764 | LR | - |
| | | General Hospital | | | | | | | | | | | | | 4: Dense | | | | |
| Tomasev et al. (2019)[93] | Acute kidney injury in 48 hours prediction | US Department of Veterans Affairs | 703,782 patients | EHR | Y | Y | Y | Y | Y | Y | - | Temporal matrix | Six hours binary occurrence | Labeled as "missing" | 1: Input<br>2: Embedding<br>3: GRU<br>4: Dense | temporal | AUC=0.921 | - | - |
| Ge et al. (2018)[38] | Mortality prediction | Asan Medical Center | 4,896 patients | EHR | Y | Y | Y | - | - | - | - | Temporal matrix | Hourly mean | "Special" category imputation | 1: Input<br>2 to k+1: *k* LSTMs for each feature<br>k+2: Dense | static | AUC=0.761 | LR | Aggregating weights of the neurons to obtain the importance |
| Bekhet et al. (2019)[50] | Heart failure prediction | Cerner Health Facts | 150,000 patients | EHR | - | - | - | Y | Y | Y | - | Multi-hot vector | - | - | RETAIN | static | AUC=0.822 | LR | - |
| Barbieri et al. (2020)[60] | Readmission in 30 days prediction | MIMIC-III | 33,150 patients | EHR | Y | - | Y | Y | Y | Y | - | Multi-hot vector | - | - | 1: Input<br>2: Embedding<br>3: Bi-GRU with attention<br>4: Dense | static | AUC=0.748 | LR, RNN | Aggregating weights of the neurons to obtain the importance |
| Qiu et al. (2019)[94] | Total joint replacement surgery risk prediction | MarketScan | 535,499 patients | AC | - | - | Y | Y | Y | Y | Revenue code and pace of service | Multi-hot vector | - | - | 1: Input<br>2: Embedding with Skip-Gram<br>3: LSTM<br>4: Dense | static | AUC=0.820 | LR, RF, CNN | - |
| Mohammadi et al. (2019)[88] | Risk of uncontrolled hypertension prediction | Partners Healthcare | 17,416 patients | EHR | Y | Y | Y | - | Y | Y | - | Multi-hot vector | - | - | 1: Input<br>2: Embedding with Skip-Gram<br>3: LSTM<br>4: Dense | static | AUC=0.719 | LR | Aggregating weights of the neurons to obtain the importance |
| Park et al. (2019)[79] | Cardio-cerebrovascular prediction | Korean National Health Insurance Service | 74,535 patients | AC | - | - | Y | - | Y | Y | - | Multi-hot vector | - | - | 1: Input<br>2: Embedding<br>3: LSTM<br>4: Dense | static | F-1=0.772 | LR, SVM, DT, RF, MLP | - |
| Caicedo-Torres et al. (2019)[39] | Mortality prediction | MIMIC-III | 22,413 patients | EHR | Y | Y | Y | - | - | - | - | Temporal matrix | Hourly mean | Forward imputation | 1: Input<br>2: Embedding<br>3: CNN<br>4: Dense | static | AUC=0.873 | - | Feature Shapley values + Heatmap of a patient |
| Sha et al. (2017)[2] | Mortality prediction | MIMIC-III | 7,537 patients | EHR | - | - | - | - | Y | - | - | Multi-hot vector | - | - | 1: Input<br>2: Embedding with Word2Vec<br>3: Bi-GRU with code | static | F-1=0.576 | LR, SVM | Temporal visualization of a patient visit records generated by |
| | | | | | | | | | | | | | | | level attention mechanism<br>4: Bi-GRU with visit level attention mechanism<br>5: Dense | | | | attention mechanism + Population level feature importance based on attention mechanism |
| Min et al. (2019)[1] | Readmission in 30 days prediction | Geisinger Health System | 27,138 patients | AC | - | - | Y | Y | Y | Y | Location | Temporal matrix | Monthly binary occurrence | - | 1: Input<br>2: GRU<br>3: Dense | static | AUC=0.65 | LR, RF, SVM, MLP, DT, CNN, LSTM | - |
| Ma et al. (2017)[69] | Diagnosis prediction | Two Insurance Claims Dataset | 147,810 patient | AC | - | - | - | Y | Y | - | - | Multi-hot vector | - | - | Dipole :<br><br>1: Input<br>2: Embedding<br>3: Bi-GRU with visit level attention mechanism<br>4: Dense | temporal | Accuracy=0.481 | RNN, RETAIN | Temporal visualization of a patient visit records generated by attention mechanism + Population level feature importance based on attention mechanism |
| Ma et al. (2018)[70] | Diagnosis prediction | A Medicaid Database and MIMIC-III | 17,584 and 99,159 patients | EHR and AC | - | - | - | - | Y | - | Ancestor of medical codes | Multi-hot vector | - | - | KAME:<br><br>1: Input<br>2: Embedding of medical codes<br>3: Embedding of ancestors of medical codes<br>4: GRU on layer 2<br>5: Softmax with attention<br>6: Dense | temporal | Accuracy@30=0.915<br>Accuracy@30=0.847 | RNN, GRAM, Dipole | Attention weight of each diagnosis code for predicting each diagnosis category |
| Ashfaq et al. (2019)[61] | Readmission in 30 days prediction | A hospital in Sweden | 7,500 patients | EHR | - | Y | Y | Y | Y | Y | - | Bag of words | - | - | 1: Input<br>2: LSTM<br>3: Dense | static | AUC=0.770 | - | - |
| Choi et al. (2016)[22] | Heart failure in 18 months prediction | Sutter Palo Alto Medical Foundation | 263,683 patients | EHR | - | - | - | Y | Y | Y | - | Multi-hot vector | - | - | RETAIN :<br><br>1: Input<br>2: Embedding | temporal | AUC=0.871 | LR, MLP, GRU | Temporal visualization of a patient visit records |

| | | | | | | | | | | | | | | | | | | |
|---|---|---|---|---|---|---|---|---|---|---|---|---|---|---|---|---|---|---|
| | | | | | | | | | | | | | | 3: Reversed GRU with attention mechanism<br>4: Reversed GRU with attention mechanism<br>5: Dense | | | | generated by attention mechanism |
| Harutyunyan et al. (2019)[12] | Mortality, length of stay, phenotyping, decompensation Prediction | MIMIC-III | 33,798 patients | EHR | Y | Y | - | - | - | - | - | Temporal matrix | Last hourly measure | Last or manual imputation | 1: Input<br>2: f Bi-LSTMs<br>3+i: LSTM<br>4+i: Dense | static and temporal | AUC=0.870<br>MAD=0.451<br>AUC=0.776<br>AUC=0.911 | LR, LSTM | - |
| Pham et al. (2017)[71] | Diagnosis prediction for mental health and diabetes | An Australian Hospital | 13,300 patients | EHR | - | - | - | Y | Y | Y | Time since last visit and admission type | Multi-hot vector on code groups | - | - | 1: Input<br>2: Embedding<br>3: Modified LSTM with decay and admission type with attention mechanism<br>4: Dense | temporal | Precision@1=52.7<br>Precision@1=66.2 | SVM, RF, RNN, LSTM | - |
| An et al. (2019)[80] | Risk of cardiovascular disease prediction | Xiangya Medical Big Data Project of Central South University | 146,296 patients | EHR | - | Y | Y | - | Y | - | - | Multi-hot vector | - | - | 1: Input<br>2 to 4: Bi-LSTMs with attention mechanism<br>5: MLP for demographic<br>6: Dense | static | F-1=0.681 | LR, SVM, RF, RNN, LSTM, Deepr, Dipole | Population level feature importance based on attention mechanism |
| Lee et al. (2019)[72] | Diagnosis prediction | MIMIC-III | 21,897 patients | EHR | - | Y | - | Y | - | Y | - | Multi-hot vector | - | - | 1: Input<br>2: Embedding<br>3: LSTM<br>4: Dense | temporal | AUC = 0.816 | LR | - |
| Rajkomar et a. (2018)[13] | Mortality, readmission, diagnosis and length of stay prediction | EHR data from the University of California, San Francisco and the University of Chicago Medicine | 114,003 patients | EHR | - | Y | Y | Y | Y | Y | Time since last visit | Multi-hot vector | Hourly mean | - | 1: Input<br>2: Embedding<br>3: LSTM with attention mechanism<br>4: Dense | static and temporal | AUC= 0.94<br>AUC=0.76<br>AUC=0.86<br>AUC=0.90 | - | Attribution mechanisms to highlight, for each patient, the data elements that influenced their predictions |
| Maragatham et al. (2019)[51] | Heart failure in 6 months prediction | An EHR Database | 34,538 patients | EHR | Y | Y | Y | Y | Y | Y | - | Multi-hot vector | - | - | 1: Input<br>2: Embedding with Skip-Gram<br>3: LSTM<br>4: Dense | static | AUC =0.894 | LR, KNNN, SVM, MLP | - |
| Lee et al. (2020) [73] | Diagnosis prediction | MIMIC-III | 46,518 patients | EHR | - | - | - | - | Y | - | - | Graph | - | - | 1: Input<br>2: GCNN<br>3: GCNN<br>4: LSTM<br>5: Dense | temporal | Rec@10=0.285<br>Prec@10=0.378 | Doctor AI | - |
| Ge et al. (2019)[95] | Post-stroke pneumonia prediction | Second Affiliated Hospital of Nanchang University in China | 13,930 patients | EHR | - | Y | Y | - | Y | Y | - | Multi-hot vector | - | - | RETAIN | static | AUC=0.928 | LR, SVM, XGBoost, MLP, RNN | Temporal visualization of a patient visit records generated by attention mechanism |
| Reddy et al. (2018)[62] | Readmission in 30 days for lupus patients prediction | Cerner HealthFacts EMR | 9,457 patients | EHR | - | - | Y | - | Y | - | Admission info | Multi-hot vector | - | - | 1: Input<br>2: Embedding<br>3: LSTM<br>4: Dense | temporal | AUC=0.700 | LR, MLP | - |
| Zhang et al. (2020)[40] | Mortality, readmission and length of stay prediction | MIMIC-III | 39,429 admissions | EHR | Y | Y | Y | - | - | - | Admission type | Temporal matrix | Hourly mean | Zero imputation | 1: Input<br>2: LSTM<br>3.-LSTM<br>4: Dense | static and temporal | AUC= 0.871<br>AUC=0.674<br>AUC=0.784 | LR, RF, CNN | - |
| Ye et al. (2020) [89] | Hypertension treatment prediction | GE Centricity Electronic Medical Record Research Database | 245,499 patients | EHR | Y | Y | Y | - | Y | Y | Lifestyle | Multi-hot vector | Monthly aggregation | - | 1: Input<br>2: Embedding<br>3: LSTM<br>4: Dense | temporal | F-1 = 0.961 | - | - |
| Yu et al. (2020)[45] | Mortality prediction | MIMIC-III | 22,049 admissions | EHR | Y | Y | - | - | Y | - | - | Multi-hot vector | - | - | 1: Input<br>2: Embedding with BoW<br>3: Bi-LSTM<br>4: Dense | static | AUC = 0.885 | - | - |
| Zhang et al. (2019)[84] | Septic shock prediction | Christiana Care Health System | 12,980 visits | EHR | Y | Y | - | - | - | - | - | Temporal matrix | Hourly mean | Zero imputation | 1: Input<br>2: Embedding<br>3: LSTM<br>4: Dense | temporal | AUC>0.774 | - | - |
| Zhang et al. (2019)[52] | Heart failure prediction | MIMIC-III | 1,700 patients | EHR | - | - | - | - | Y | - | Time since last visit and time to the index date | Multi-hot vector | - | - | 1: Input<br>2: Embedding for diagnosis and time | temporal | AUC=0.737 | LR, RF, SVM, GRU, LSTM, REATIN, | Temporal visualization of a patient visit records generated by |
| | | | | | | | | | | | | | | | 3: LSTM for diagnosis and time<br>4: LSTM for time and diagnosis with knowledge attention mechanism<br>5: Dense | | | GRAM, KAME | attention mechanism + Population level feature importance based on average attention |
| Wickramaratne et al. (2020)[85] | Sepsis detection | Physionet 2019 Challenge Database | 40,336 patients | EHR | Y | Y | Y | - | - | - | - | Temporal Matrix | Hourly mean | - | 1: Input<br>2: Bi-GRU<br>3: Dense | static | AUC=0.97 | - | - |
| Shickel et al. (2019)[41] | Mortality prediction | University of Florida and MIMIC-III | 27,660 and 35,993 patients | EHR | Y | Y | - | - | - | - | - | Temporal matrix | Hourly mean | Forward imputation | 1: Input<br>2: GRU with attention mechanism on time<br>3: Dense | static | AUC>0.900 | - | Temporal visualization of a patient visit records generated by attention mechanism |
| Choi et al. (2017)[53] | Diagnosis prediction and heart failure prediction | Sutter Palo Alto Medical Foundation primary care patients and MIMIC III | 30,727 and 7,499 patients | EHR | - | - | - | - | Y | - | - | Multi-hot vector | - | - | 1: Input<br>2: Embedding with medical ontology<br>3: GRU<br>4: Dense | static and temporal | Accuracy@20=0.264<br>AUC=0.844 | GRU | - |
| Razavian et al. (2016)[96] | Renal disease prediction | An EHR Database | 5,484 patients | EHR | - | Y | Y | - | Y | Y | - | Temporal matrix | Monthly mean | - | 1: Input<br>2: CNN<br>3: CNN<br>4: Dense | temporal | AUC=0.917 | LR, LSTM | - |
| Purushotham et al. (2018)[42] | Mortality, length of stay, phenotyping, decompensation prediction | MIMIC-III | 35,627 patients | EHR | Y | Y | Y | - | - | - | - | Temporal matrix | Hourly mean | Forward and backward imputation | 1: Input<br>2: GRU on temporal<br>3: MLP on static<br>4: Dense | static and temporal | AUC=0.873<br>MSE=369241<br>AUC=0.777<br>AUC=0.871 | LR, MLP, RF, DT, NB, RNN | - |
| Gupta et al. (2020)[43] | Mortality prediction, diagnosis prediction | MIMIC-III | 42,276 patients | EHR | Y | Y | Y | - | - | - | - | Temporal Matrix | - | - | 1: Input<br>2: GRU<br>3: GRU<br>4: Dense | static and temporal | AUC=0.852<br>AUC=0.822 | LR | Aggregating weights of the neurons to obtain the importance |
| Ma et al. (2018)[54] | Heart failure in 12 months prediction | SNOW dataset | 2,268 patients | EHR | - | - | - | Y | Y | Y | - | Multi-hot vector | - | - | 1: Input<br>2: Embedding with Word2Vec<br>3: Bi-GRU with | temporal | AUC=0.728 | LR, RETAIN | Temporal visualization of a patient visit records generated by attention mechanism |

| | | | | | | | | | | | | | | | | | | |
|---|---|---|---|---|---|---|---|---|---|---|---|---|---|---|---|---|---|---|
| | | | | | | | | | | | | | | attention mechanism<br>4: CNN with time awareness<br>5: Dense | | | | |
| Zheng et al. (2017)[91] | Illness severity prediction | National University Hospital in Singapore | 2,740 patients | EHR | - | Y | Y | - | - | - | Time interval between visits and between lab tests | Multi-hot vector | - | - | 1: Input<br>2: Embedding<br>3: Modified GRU with visit and feature decay<br>4: Dense | temporal | MSE=4.146 | - | - |
| Lipton et al. (2016)[74] | Diagnosis prediction | Children's Hospital LA | 10,401 admissions | EHR | Y | Y | - | - | - | - | - | Temporal matrix | Hourly mean | Zero imputation | 1: Input<br>2: LSTM with target replication<br>3: LSTM with target replication<br>4: Dense | temporal | AUC=0.873<br>F-1=0.304 | LR, MLP | - |
| Morid et al. (2020)[98] | Cost prediction | University of Utah Health Plans | 91,000 patients | AC | - | - | - | Y | Y | Y | Previous cost and visit type | Temporal matrix | Monthly aggregation measures | - | 1: Input<br>2: CNN<br>3: CNN<br>4: Dense | static | MAPE=1.82 | - | - |
| Solares et al. (2020)[56] | Hearth failure and emergency admission in 6 months prediction | UK National Health Service | 63,808 patients | EHR | - | - | Y | - | Y | Y | Time since last visit | Multi-hot vector | - | - | RETAIN | temporal | AUC=0.950<br>AUC=0.847 | LR, Deepr | - |
| Bai et al. (2018)[75] | Diagnosis prediction | SEER-Medicare Database | 45,104 patients | AC | - | - | - | Y | Y | - | Time since last visit | Multi-hot vector | - | - | 1: Input<br>2: Embedding with attention based on time interval<br>3: Bi-LSTM<br>4: Dense | temporal | F-1=0.530 | RNN, Dipole | Temporal visualization of a patient visit records generated by attention mechanism |
| Choi et al. (2018)[55] | Heart failure prediction | Sutter Palo Alto Medical Foundation | 30,764 patients | EHR | - | - | - | - | Y | Y | - | Multi-hot vector | - | - | 1: Input<br>2: Embedding layer considering Rx as a sub category of Dx<br>3: GRU<br>4: Dense | static | AUC=0.4787 | LR, GRAM | - |
| Rebane et al. (2019)[97] | Adverse drug event prediction | HealthBank at Stockholm University | 1,872 patients | EHR | - | - | - | - | Y | Y | - | Multi-hot vector | - | - | RETAIN | temporal | AUC=0.789 | - | Temporal visualization of a patient visit |
| | | | | | | | | | | | | | | | | | | | records generated by attention mechanism |
| Baker et al. (2020)[44] | Mortality prediction | MIMIC-III | 22,483 patients | EHR | Y | - | Y | - | - | - | - | Temporal matrix | Monthly aggregation measures | Constant imputation | 1: Input<br>2: CNN<br>3: CNN<br>4: Bi-LSTM<br>5: Dense | static | AUC=0.884 | - | - |
| Fagerström et al. (2019)[87] | Septic shock prediction | MIMIC-III | 50,373 patients | EHR | Y | Y | - | - | - | - | - | Temporal matrix | Hourly mean | Last value imputation | 1: Input<br>2 to 5: LSTM<br>6: Dense | static | AUC=0.83 | - | - |
| Svenson et al. (2020)[86] | Sepsis deterioration prediction | MIMIC-III | 5,784 admissions | EHR | Y | Y | - | - | - | - | - | Temporal matrix | Last hourly measure | Last or manual imputation | 1: Input<br>2 to i: LSTMs<br>3+i: LSTM<br>4+i: Dense | temporal | AUC=0.846 | - | - |
| Sue et al. (2017)[92] | Illness severity prediction | University Hospital of Catanzaro | 9,704 patients | EHR | - | Y | - | - | - | - | - | Temporal matrix | Hourly mean | Last imputation | 1: Input<br>2: Embedding layer with attention based<br>3: GRU<br>4: Softmax | temporal | Accuracy=0.845 | LR | Temporal visualization of a patient visit records generated by attention mechanism |
| Liu et al. (2020)[76] | Diagnosis prediction | MIMIC-III | 22,483 patients | EHR | Y | Y | - | - | - | - | - | Multi-hot vector | - | - | 1: Input<br>2: Embedding<br>3: Modified GRU with visit and feature decay<br>4: Softmax | temporal | AUC=0.90 | LR, SVM, RF, Doctor AI | - |
| Zhang et al. (2020)[77] | Diagnosis prediction | MIMIC-III | 7,499 patients | EHR | - | - | - | - | Y | - | - | Multi-hot vector | - | - | 1: Input<br>2: Embedding with medical ontology<br>3: GRU<br>4: Softmax | temporal | Accuracy@20=0.390 | GRAM | - |
| Xiang et al. (2020)[99] | Asthma exacerbation prediction | Cerner Health Facts database | 31,433 patients | EHR | - | - | - | - | - | Y | Y | Multi-hot vector | - | - | 1: Input<br>2: Embedding<br>3: LSTM with code level attention mechanism<br>4: LSTM with visit level attention mechanism<br>5: Softmax | temporal | AUC=0.700 | LR, MLP, LSTM , RETAIN | Temporal visualization of a patient visit records generated by attention mechanism |
| Thorsen-Meyer et al. (2020)[90] | Decompensation prediction | Danish National | 12,616 patients | EHR | Y | Y | Y | - | - | - | - | Temporal matrix | Hourly median | Last imputation | 1: Input<br>2: LSTM | temporal | AUC=0.820 | - | Feature Shapley values |
| | | Patient Registry | | | | | | | | | | | | | 3: Softmax | | | | |
| Qiao et al. (2020)[78] | Diagnosis prediction | MIMIC-III | 7,537 patients | EHR | Y | Y | - | - | - | Y | Y | Multi-hot vector | - | - | 1: Input<br>2: Embedding for medical codes<br>3: Time-series feature extractor with HCTA<br>4: Bi-GRU with attention<br>5: Softmax | temporal | Recall@30=0.642 | Dipole, RETAIN, Doctor AI | - |